\documentclass[10pt,twocolumn,letterpaper]{article}

\usepackage[pagenumbers]{cvpr} % To force page numbers, e.g. for an arXiv version

\usepackage[dvipsnames]{xcolor}

\usepackage{multirow}
\usepackage{float}
\usepackage{pifont}% http://ctan.org/pkg/pifont
\usepackage{caption, subcaption}
\usepackage{listings}
\usepackage{array}
\usepackage{graphicx}

\newcommand{\xmark}{\ding{55}}%

\newcommand{\ceil}[1]{\left\lceil #1 \right\rceil}
\newcolumntype{P}[1]{>{\centering\arraybackslash}p{#1}}

\usepackage{color, colortbl}
\definecolor{Gray}{gray}{0.9}
\definecolor{LightBlue}{rgb}{0.68, 0.85, 0.9}
\definecolor{LightGreen}{rgb}{0.34, 0.91, 0.48}

\usepackage{xcolor}

\definecolor{codegreen}{rgb}{0,0.6,0}
\definecolor{codegray}{rgb}{0.5,0.5,0.5}
\definecolor{codepurple}{rgb}{0.58,0,0.82}
\definecolor{backcolour}{rgb}{0.95,0.95,0.92}

\lstdefinestyle{mystyle}{
    backgroundcolor=\color{backcolour},
    commentstyle=\color{codegreen},
    keywordstyle=\color{magenta},
    numberstyle=\tiny\color{codegray},
    stringstyle=\color{codepurple},
    basicstyle=\ttfamily\footnotesize,
    breakatwhitespace=false,
    breaklines=true,
    captionpos=b,
    keepspaces=true,
    numbers=left,
    numbersep=5pt,
    showspaces=false,
    showstringspaces=false,
    showtabs=false,
    tabsize=2
}

\definecolor{cvprblue}{rgb}{0.21,0.49,0.74}
\usepackage[pagebackref,breaklinks,colorlinks,citecolor=cvprblue]{hyperref}
\usepackage{caption}
\def\paperID{2871} % *** Enter the Paper ID here
\def\confName{CVPR}
\def\confYear{2024}

\title{Just Add \scalebox{1.35}{$\boldsymbol{\pi}$}! \underline{P}ose \underline{I}nduced Video Transformers for \\ Understanding Activities of Daily Living}

\author{Dominick Reilly\\
UNC Charlotte\\
{\tt\small dreilly1@charlotte.edu}
\and
Srijan Das\\
UNC Charlotte\\
{\tt\small sdas24@charlotte.edu}
}

\begin{document}
\twocolumn[{%
\renewcommand\twocolumn[1][]{#1}%
\maketitle
% \includegraphics[width=.195\textwidth]
% {figure_dataset/scene_2_Crossroad_campus_30Sec_droneView_01_000332.PNG}
% \hfill
% \includegraphics[width=.195\textwidth]{figure_dataset/scene_2_Crossroad_campus_30Sec_groundView_01_000332.PNG}
% \hfill
\includegraphics[width=.4925\textwidth]{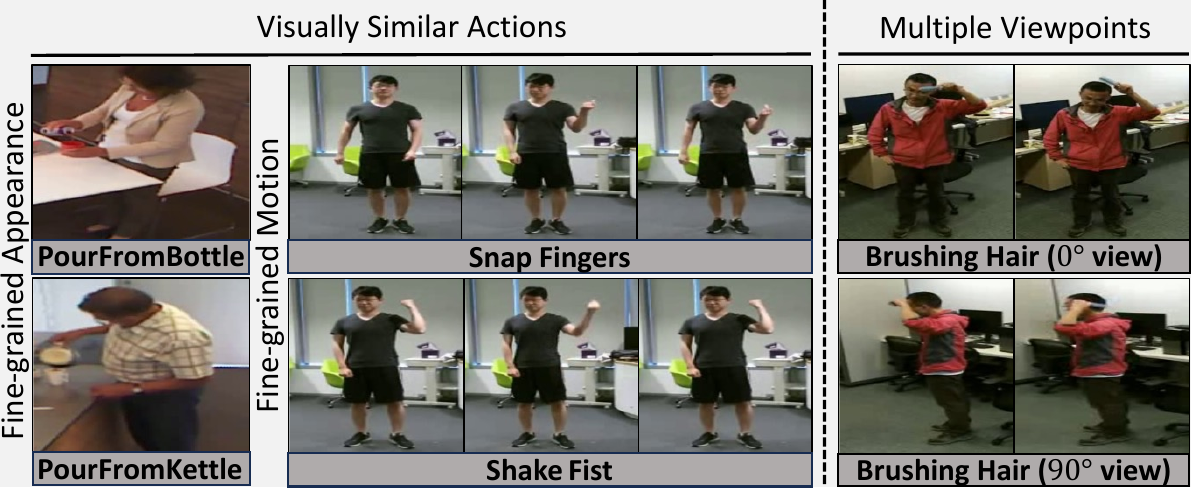}
\hfill
\includegraphics[width=.4925\textwidth]{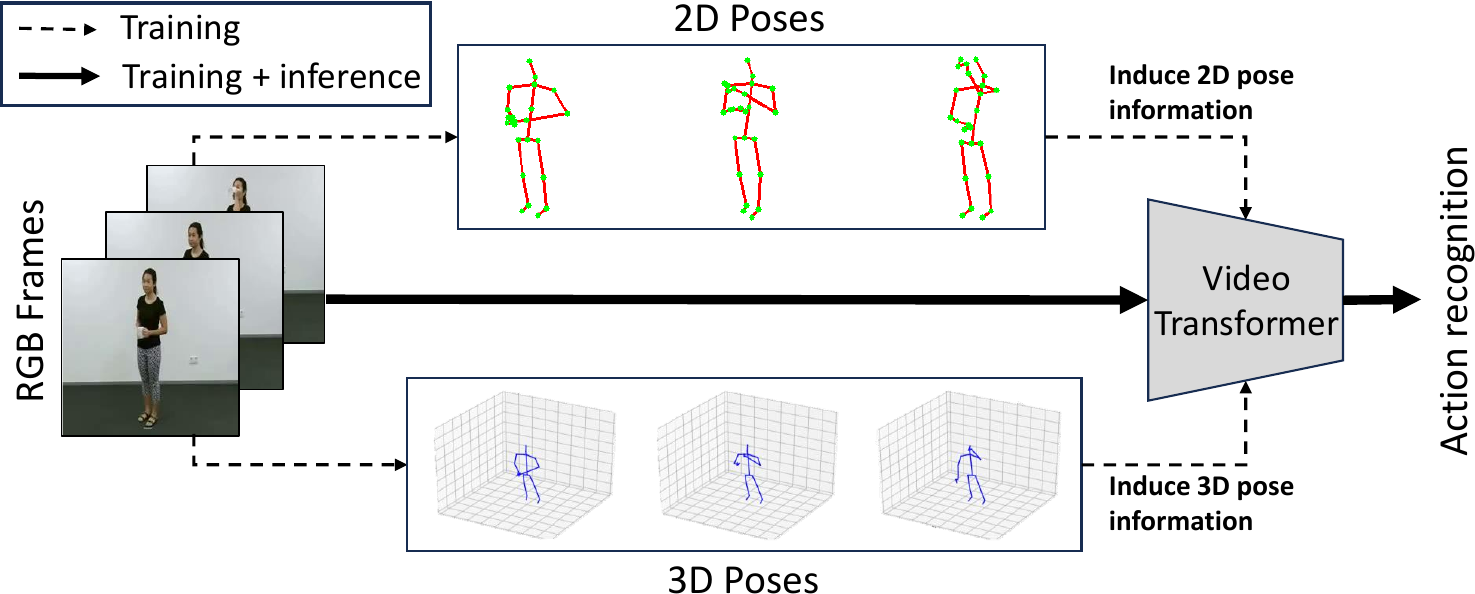}
\captionof{figure}{\textbf{Left:} we illustrate the challenges of Activities of Daily Living. Notice the visual similarity in the action pairs (\textit{PourFromBottle}, \textit{PourFromKettle}) and (\textit{Snap Fingers}, \textit{Shake Fist}). Also notice the significant change in appearance when viewing \textit{Brushing Hair} from different angles. \textbf{Right:} we present an overview of our proposed approach to address these challenges. We induce human pose information into the representations learned by video transformers. This induction of information is only required during training.}
% \begin{figure*}[!h]
%     \centering
%     \begin{subfigure}{0.52\textwidth}
%         \includegraphics[width=\textwidth]{content/intro-fig_cropped_v3.pdf}
%         \caption{The challenges of Activities of Daily Living. The challenges of Activities of Daily Living. The challenges of Activities of Daily Living. }
%         \label{fig:intro_diagramb}
%     \end{subfigure}
%     \hfill
%     \begin{subfigure}{0.46\textwidth}
%         \includegraphics[width=\textwidth]{content/intro-fig_cropped.pdf}
%         \caption{An overview of our approach. 2D and 3D skeleton information is incorporated into an RGB model during training.}
%         \label{fig:intro_diagrama}
%     \end{subfigure}
%     \caption{An overview of our approach and the challenges of activities of daily living}
% % \end{figure*}
\label{fig:combined_figures}
\vspace{1em}
}]

%%
%% We modified "\def\endabstract" in cvpr.sty
%%
\begin{abstract}
\vspace*{-2em}

Video transformers have become the de facto standard for human action recognition, yet their exclusive reliance on the RGB modality still limits their adoption in certain domains. One such domain is Activities of Daily Living (ADL), where RGB alone is not sufficient to distinguish between visually similar actions, or actions observed from multiple viewpoints. To facilitate the adoption of video transformers for ADL, we hypothesize that the augmentation of RGB with human pose information, known for its sensitivity to fine-grained motion and multiple viewpoints, is essential. Consequently, we introduce the first Pose Induced Video Transformer: \textbf{PI-ViT} (or {\Large$\pi$}-ViT), a novel approach that augments the RGB representations learned by video transformers with 2D and 3D pose information. The key elements of {\Large$\pi$}-ViT are two plug-in modules, 2D Skeleton Induction Module and 3D Skeleton Induction Module, that are responsible for inducing 2D and 3D pose information into the RGB representations. These modules operate by performing pose-aware auxiliary tasks, a design choice that allows {\Large$\pi$}-ViT to discard the modules during inference. 
Notably, {\Large$\pi$}-ViT achieves the state-of-the-art performance on three prominent ADL datasets, encompassing both real-world and large-scale RGB-D datasets, without requiring poses or additional computational overhead at inference. We release code and models at \href{https://github.com/dominickrei/pi-vit/}{https://github.com/dominickrei/pi-vit/}.
\end{abstract}

\section{Introduction}
Recently, the task of monitoring Activities of Daily Living (ADL) has gained prominence as it enables many applications, such as advanced security systems or assisting the elderly.
% especially in the development of intelligent systems designed to assist the elderly.
Historically, the learning of ADL representations has borrowed heavily from the action recognition literature, where advanced vision models are trained primarily on internet-sourced videos~\cite{kinetics, kinetics600, ucf, kuehne2011hmdb} such as sports, YouTube, and movie clips.
These models, however, are predominantly appearance-based, aligning actions strongly with their scenes~\cite{actionrecognition_without_human_ECCVW16}, and do not adequately capture the intrinsic challenges of ADL. 
In Figure~\ref{fig:combined_figures}, we present the challenges in ADL which broadly includes the presence of visually similar actions and actions captured from different camera views. %Firstly, visually similar actions in ADL contain \textit{fine-grained} details. These actions must be distinguished from one another through subtle cues in appearance or motion. As an example, to distinguish actions "drink from bottle" and "drink from cup", the model must understand the subtle differences in appearance of "bottle" and "cup". On the other hand, to distinguish between the actions "snap fingers" and "shake fist", the model must understand the subtle differences in motion patterns between them. Secondly, the actors of ADL can be observed from multiple viewpoints, thus requiring the model to have view-agnostic representations that can recognize actions regardless of viewpoint.
Firstly, ADL involves visually similar but \textit{fine-grained} actions, distinguishable through subtle appearance or motion cues. %For instance, actions like "drink from bottle" vs "drink from cup" can be disambiguated by the subtle differences in appearance of "bottle" and "cup", and  "snap fingers" vs "shake fist" can be distinguished by subtle differences in motion patterns. 
Secondly, the actors of ADL can be observed from multiple viewpoints, thus requiring the learning of view-agnostic representations that can recognize actions regardless of camera viewpoint. 

% Action representation learning has undergone a paradigm shift with the emergence of vision transformers. Echoing trends in other vision tasks~\cite{clip_representation, carion2020detr, xie2021segformer}, these transformers~\cite{attention} have garnered acclaim for their effectiveness and robustness across different modalities~\cite{dosovitskiy2020vit, BERTnlp, wav2vec2_speech}.
Video representation learning has undergone a paradigm shift with the emergence of video transformers~\cite{timesformer, vivit, liu2021videoswin}. However, despite their effectiveness in learning action representations, they predominantly rely only on the RGB modality. These unimodal RGB representations are insufficient to capture the fine-grained details in the videos and are sensitive to changes in viewpoint. Thus, these visual models are suboptimal for ADL and strongly warrant the need for research directed towards multi-modal representation learning.

Furthermore, huge advancements have been made in the field of skeleton based action recognition~\cite{stgcn, 2sagcn2019cvpr, 3mformer_CVPR23, zhou2022hyperformer} using 3D human poses, which are inherently viewpoint-agnostic and offer key positional information for modeling human motion. 
They are effective for some ADL challenges but they cannot encode appearance information. 
Therefore, a natural idea is to combine RGB and poses~\cite{das2020vpn, PoseC3D_CVPR22, Ketul_ContrastiveMultiview_WACV23, Ahn2022_STARTransformerAR_WACV23, Kim_3DDeformableAttnForActionRec_ICCV23}. However, these multi-modal methods require depth sensors for obtaining 3D poses or they incur high latency due to the computational demands of estimating 3D poses from RGB~\cite{lcrnet_new, videopose3d}. 
This brings us to the main question: \textit{What is the best strategy to combine RGB and Poses without compromising model latency?}

\begin{figure}[t]
    \centering
    \includegraphics[width=0.92\linewidth]{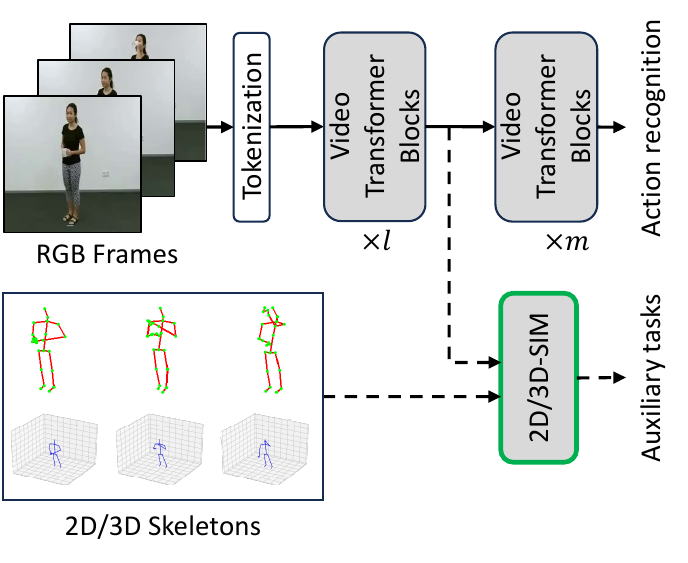}
    \caption{\textbf{An overview of our Pose Induced Video Transformer ({\Large$\pi$}-ViT).} During training (indicated by dashed lines), the video transformer incorporates 2D-SIM and 3D-SIM, which process skeletons and the intermediate visual representations. During inference the video transformer is used independently.}
    \label{fig:pi-vit-highlevel}
    \vspace{-1.8em}
\end{figure}

We observe that crucial appearance cues for distinguishing visually similar actions are localized in the RGB regions corresponding to the salient human skeleton joints. Additionally, the temporal evolution of 3D skeletons effectively captures fine-grained motion and is viewpoint-agnostic. We hypothesize that infusing these inherent properties of 3D skeletons into video transformers will optimize the RGB representations to address the challenges of ADL.

To this end, we introduce the \textbf{Pose Induced Video Transformer}, dubbed as PI-ViT or {\Large$\pi$}-ViT. It utilizes both 2D and 3D skeletons to infuse complementary information into the visual token representations learned by video transformers (see Figure~\ref{fig:combined_figures}). 
{\Large$\pi$}-ViT is composed of two novel plug-in modules: 2D Skeleton Induction Module (2D-SIM) and 3D Skeleton Induction Module (3D-SIM). 
2D-SIM leverages 2D skeletons to perform an auxiliary task of mapping the skeleton joints and visual tokens to provide extra supervision to the RGB regions containing the relevant skeleton joints involved in an action. This task refines the RGB representations and enforces the video transformer to discriminate actions with fine-grained appearance. 
Conversely, 3D-SIM utilizes 3D skeletons to address the challenges of fine-grained motion and multiple viewpoints. To realize this objective, 3D-SIM performs an auxiliary feature alignment task, refining the RGB representations by integrating an optimized 3D skeleton representation for action classification. 
These modules, 2D-SIM and 3D-SIM, can be inserted after any layer of the existing video transformer.
What sets {\Large$\pi$}-ViT apart is its pose induction through auxiliary tasks performed by the modules. This not only enforces that the video transformer learn a pose-augmented RGB representation but also allows for the removal of these modules during inference, eliminating the need for poses at inference time. 

\begin{figure*}[t]
    \centering
    \includegraphics[width=0.9\textwidth]{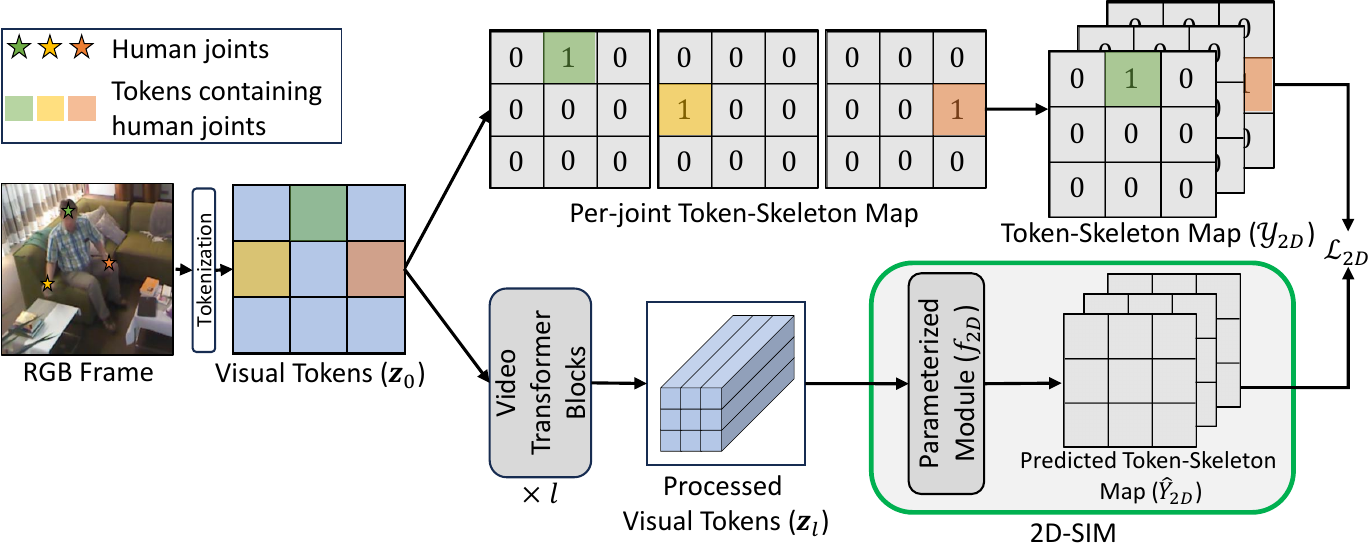}
    \caption{\textbf{An overview of the 2D Skeleton Induction Module (2D-SIM).} We only visualize a single frame from the video for clarity. A token-skeleton map is constructed that indicates the presence of human skeleton joints in the RGB regions corresponding to the visual token. 2D-SIM refines the visual tokens from the video transformer, and uses them to predict the token-skeleton map.}
    \label{fig:2dsim-detailed}
    \vspace{-1em}
\end{figure*}

\noindent We summarize the key contributions of our work as:
\begin{itemize}
    \item The introduction of the first Pose Induced Video Transformer, {\Large$\pi$}-ViT, that leverages both 2D and 3D Poses for enhancing video representation learning on ADL.
    \item{\Large$\pi$}-ViT incorporates two novel plug-in modules, 2D-SIM and 3D-SIM, that are designed to address the challenges of ADL. These modules perform distinct pose-aware auxiliary tasks that enable video transformers to learn fine-grained and view-invariant representations.
    \item{\Large$\pi$}-ViT's efficacy is demonstrated through superior performance on the real-world ADL dataset Toyota-Smarthome, and the largest RGB-D human action recognition datasets: NTU120 and NTU60. {\Large$\pi$}-ViT achieves state-of-the-art results without requiring poses, or additional computational overhead, during inference.
\end{itemize}

\section{Preliminaries: Video Transformers}

\label{sec:preliminaries}
In this section, we provide a brief description of the working principle of existing video transformers~\cite{timesformer, liu2021videoswin, vivit, mvit1}.
Given a video $V$ with the shape $T \times H \times W \times 3$, video transformers decompose $V$ into disjoint spatio-temporal patches, each of size $\tau \times p \times p$, where $\tau=1$ is similar to image patches~\cite{timesformer} and $\tau>1$ is similar to the tublets employed in~\cite{vivit, liu2021videoswin, mvit1}. Then, these patches are tokenized via a linear projection that projects the patches to the shape $T_{v} \times S_v \times d_{v}$, where the spatial dimension $S_v = \ceil{\frac{H}{p}} \cdot \ceil{\frac{W}{p}}$ and $d_v$ is the embedding dimension of the video transformer.
% These tokens are also known as patches~\cite{timesformer, mvit1} and sometimes as tublets~\cite{vivit} or 3D patches~\cite{liu2021videoswin} if $T < \bar{T}$. In this paper, we only focus on video transformers where $T = \bar{T}$.
Spatio-temporal positional embeddings are then added to each of the tokens, enabling them to encode their location in the video. Additionally, a classification token is appended, resulting in a total of $T_{v} \cdot S_{v} + 1$ tokens. The resulting sequence of tokens $\mathbf{z}_0$ can then be processed by a series of video transformer blocks. Consequently, the output of the $l^\text{th}$ video transformer block, $\mathbf{z}_l$, can be obtained as
%%%% Set the display skip for the entire paper
\setlength{\abovedisplayskip}{3pt}
\setlength{\belowdisplayskip}{3pt}
\begin{equation}
    \mathbf{z}_l = \mathbf{z}_{l-1} + \text{ST-MHSA}(\text{LN}(\mathbf{z}_{l-1}))
\end{equation}
\begin{equation}
    \mathbf{z}_l = \mathbf{z}_l + \text{MLP}(\text{LN}(\mathbf{z}_l))
\end{equation}
where ST-MHSA denotes spatio-temporal multihead self-attention~\cite{timesformer}, LN denotes layer normalization~\cite{layernormalization_2016}, and MLP denotes multi-layer perceptron. In a video transformer composed of $L$ layers, class predictions are computed from the classification token in $\mathbf{z}_L$ via a fully connected layer. The video transformer is then trained with an entropy loss ($\mathcal{L}^{cls}_{v}$) computed using the class predictions and ground-truth.

\section{Proposed Video Transformer}

% Next, we present our Pose Enhanced Video Transformer ({\Large$\pi$}-ViT) for video representation learning, which implicitly learns RGB representations that are view-invariant and can discriminate the fine-grained and similar actions that are commonplace in ADL. This is accomplished by the introduction of 2D and 3D human skeleton information during training through the use of two modules: (1) the 2D Pose Enhancement Module (2D-SIM) and (2) the 3D Pose Enhancement Module (3D-SIM). A high-level overview of {\Large$\pi$}-ViT is shown in Figure~\ref{fig:{\Large$\pi$}-ViT-highlevel} and 2D-SIM and 3D-SIM are detailed in Figure~\ref{fig:2dSIM-detailed} and Figure~\ref{fig:3dSIM-detailed}. Both modules are implemented to perform auxiliary tasks that are only required during training and are removed during inference.
In this section, we present our Pose Induced Video Transformer ({\Large$\pi$}-ViT) which implicitly learns discriminative representations for understanding ADL videos. This is accomplished by introducing 2D and 3D human skeleton information into a vanilla video transformer through the addition of two modules: (1) the 2D Skeleton Induction Module (2D-SIM) and (2) the 3D Skeleton Induction Module (3D-SIM). 
Throughout this paper, we use the term ``pose`` to refer to the abstract configuration of the human joints, and the term ``skeleton`` to refer the low-level positional coordinates of the human joints. A high-level overview of {\Large$\pi$}-ViT is shown in Figure~\ref{fig:pi-vit-highlevel}. Both modules are implemented to perform auxiliary tasks that are only required during training, and are removed during inference, thus requiring no extra computation during inference.

% \begin{figure*}[t]
%     \begin{subfigure}{\textwidth}
%         \includegraphics[width=\textwidth]{content/2dsim-fig-a_cropped.pdf}
%         \caption{\textbf{Token-Skeleton Map Construction}. Displayed for a single frame for simplicity.}
%         \label{fig:2d-posemap-construction}
%     \end{subfigure}
%     \hfill
%     \begin{subfigure}{\textwidth}
%         \includegraphics[width=\textwidth]{content/rough_2dSIM_b.png}
%         \caption{2D-SIM. Displayed for a single frame for simplicity.}
%         \label{fig:b}
%     \end{subfigure}
%     \caption{2D-Pose Enhancement Module (2D-SIM)}
%     \label{fig:2dSIM-detailed}
% \end{figure*}

% \fcolorbox{LightGreen}{white}{3D Skeleton Induction Module (3D-SIM)}
\begin{figure*}[t]
\centering
    \includegraphics[width=0.96\textwidth]{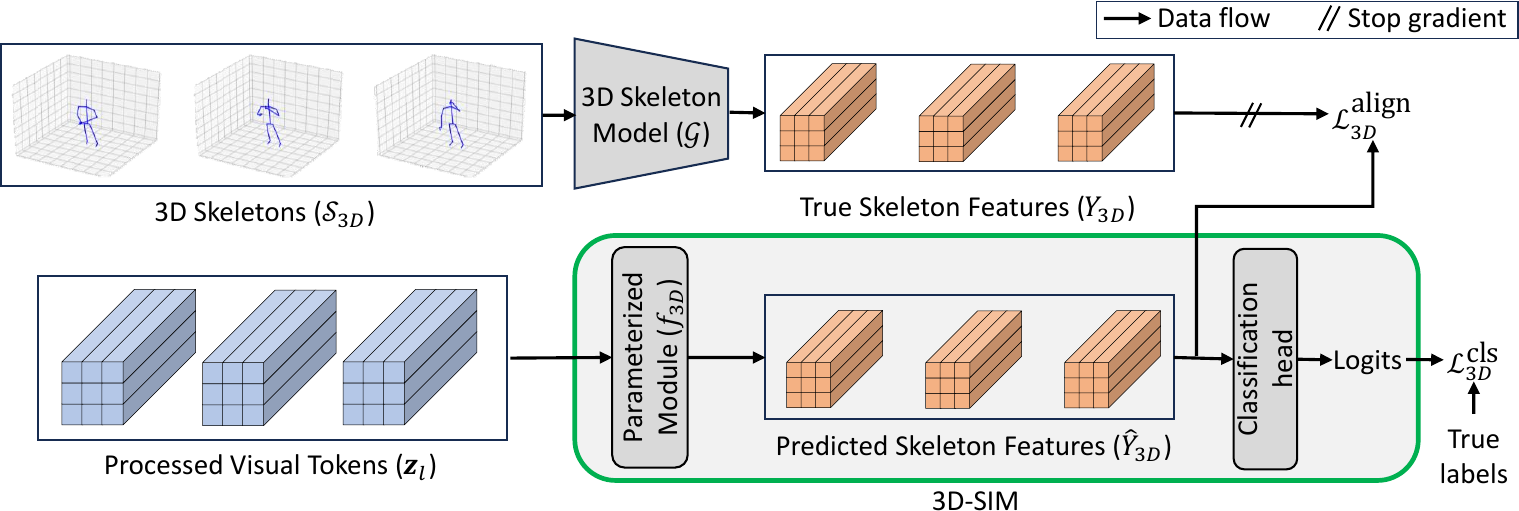}
    \caption{\textbf{An overview of the 3D Skeleton Induction Module (3D-SIM).} 3D-SIM processes visual tokens from the video transformer layer preceding it. A parameterized module transforms these tokens to obtain predicted skeleton features, which are then aligned with corresponding features generated by a pre-trained 3D skeleton model. During training, the weights of the skeleton model are not updated.}
    \label{fig:3dSIM-detailed}
    \vspace{-1em}
\end{figure*}

\subsection{2D Skeleton Induction Module (2D-SIM)}

\label{sec:2dsim}
% We first describe 2D-SIM, a plug-in module that can be inserted after any layer in an existing video transformer architecture. During the training phase, 2D-SIM's role is to refine the representations learned by the video transformer by enriching them with detailed human anatomy information from 2D skeletons. During the inference phase, 2D-SIM is removed and only the vanilla video transformer architecture is used. Specifically, 2D-SIM performs an auxiliary task that identifies the RGB regions that correspond to the locations and labels of 2D skeleton joints.
We first describe 2D-SIM, a plug-in module that can be inserted after any layer in an existing video transformer architecture, and that is designed to address the challenge of fine-grained appearance in ADL. During the training phase, the role of 2D-SIM is to refine the video representations learned by the transformer with detailed human anatomy information obtained from 2D skeletons. Specifically, 2D-SIM provides extra supervision to the RGB regions containing the specific human skeleton joints involved in an action. This extra supervision is achieved through an auxiliary task that learns the mapping between the visual tokens and the skeleton joints. At inference time, 2D-SIM is removed.

\textbf{Token-Skeleton Map.}\quad  The first step in 2D-SIM is the construction of a token-skeleton map that defines the correspondence between the RGB regions and the 2D skeleton joints. A visual illustration of this mapping can be found in Figure \ref{fig:2dsim-detailed}. The tokenization performed by the video transformer provides an elegant way to obtain distinct and discrete RGB regions. Hence, we construct a mapping from visual tokens to 2D skeleton joints. We denote the set of 2D skeleton joints for a given video $V$ as
\begin{equation}
    \mathcal{S}_{2D} = \{(t, j, x, y)\} : 1 \leq t \leq T, 1 \leq j \leq J
\end{equation}
where $J$ is the number of human joints contained in the 2D skeleton and $x, y$ indicates the spatial location of $j^\text{th}$ joint in the $t^\text{th}$ frame. Then, a binary map $\mathcal{M}$ with the shape $T \times H \times W \times J$ is computed through
\begin{equation}
    \mathcal{M}_{thwj} = 
    \begin{cases}
        1 & \textrm{if} \: (t, j, h, w) \in \mathcal{S}_{2D}\\
        0 & \textrm{otherwise}
    \end{cases}
    \label{}
\end{equation}
where each value in $\mathcal{M}$ indicates the presence or absence of a particular joint \textit{at the pixel level} of the input video. 
Since the objective is to obtain the mapping at the token level, we perform max pooling on the binary map $\mathcal{M}$ across the spatio-temporal resolution. Consequently, the token-skeleton map $\mathcal{Y}_{2D}$ is computed with a pooling kernel of size equal to $\tau \times p \times p$ as
\begin{equation}
    \mathcal{Y}_{2D} = \text{MaxPool}_{\tau \times p \times p} (\mathcal{M})
\end{equation}
Now, $\mathcal{M}$ takes the shape of $T_{v} \times S_{v} \times J$ and indicates the presence or absence of a particular joint \textit{at the token level}.

% \textbf{The Task of 2D-SIM.}\quad The goal of 2D-SIM is to enforce that the RGB representations learned by the video transformer are effective for identifying the RGB regions containing human joints, in addition to the primary task of action classification. Given the output tokens of the video transformer at the layer preceding 2D-SIM, $\mathbf{z}_l$, a joint-classifier directly predicts the presence or absence of every joint in $J$ for each token in $\mathbf{z}_l$. The joint-classifier can be implemented as any parameterized module (e.g., FC, MLP, transformer) plus a classification head. We find that a simple FC layer is sufficient and obtain the per-token joint predictions as
% \begin{equation}
%     P = \text{FC}_2(\text{FC}_1(\textbf{z}_l))
% \end{equation}
% where $\text{FC}_1$ projects from $d_{vis} \rightarrow d_b$ and $\text{FC}_2$ projects from $d_b \rightarrow J$, and $d_b$ is the bottleneck dimension\footnote{In practice $d_b=256$}. Thus, the output shape of the joint-classifier is the same as the 2D pose map, i.e., $T_{v} \times S \times J$. We compute the loss of 2D-SIM as the binary cross-entropy (BCE) loss between $P$ and $\mathcal{P}$
% \begin{equation}
%     \mathcal{L}_{2D} = BCE(\mathcal{P}, P)
% \end{equation}
\textbf{Auxiliary Task of 2D-SIM.}\quad Since the goal of 2D-SIM is to learn  representations for fine-grained appearance actions, it enforces extra supervision to the RGB regions containing human joints. This is achieved through the use of a multi-class skeleton joint classification in the video transformer. 
Given the output tokens of the video transformer at the layer preceding 2D-SIM, $\mathbf{z}_l$, 2D-SIM directly predicts the presence or absence of every joint for each token in $\mathbf{z}_l$. Thus, it is a parameterized module $f_{2D}()$ with a skeleton joint classification head. 
The token-skeleton map predictions of 2D-SIM are computed as
\begin{equation}
    \hat{Y}_{2D} = W  f_{2D}(\textbf{z}_l) + b
\end{equation}
where $W$ and $b$ are the parameters of the skeleton joint classification head. 
In practice, we find that $f_{2D}()$ implemented using a simple fully-connected layer that projects $z_l$ from $d_{v} \rightarrow d_b$ embedding space is sufficient to learn the token-skeleton mapping. 
After performing the prediction, the output shape of 2D-SIM aligns with that of the token-skeleton map, i.e., $T_{v} \times S_v \times J$. The loss for 2D-SIM is computed as the binary cross-entropy between $\mathcal{Y}_{2D}$ and $\hat{Y}_{2D}$
\begin{equation}
    \mathcal{L}_{2D} =  -\frac{1}{J} \sum_{i=1}^{J} [\mathcal{Y}_{2D}^i \log(\hat{Y}_{2D}^i) + (1 - \mathcal{Y}_{2D}^i ) \log(1 - \hat{Y}_{2D}^i)]
\end{equation}

\subsection{3D Skeleton Induction Module (3D-SIM)}

% Next we describe 3D-SIM. 3D-SIM is similar to 2D-SIM in that it is also a plug-in module that can be inserted after any layer in existing video transformer architectures; however, it plays a different role. While 2D skeletons are effective for finding correspondences between RGB and joint regions, they are not invariant to new or difficult viewpoints (\textcolor{red}{see introduction}). In contrast, 3D skeletons are viewpoint invariant.
Next we describe 3D-SIM, which shares the same modular design as 2D-SIM, allowing it to be inserted after any layer within the video transformer architecture. % Unlike its 2D counterpart, 3D-SIM is tailored to address the limitations posed by challenging viewpoints and temporally similar actions, a shortfall of 2D skeletons. 
Unlike its 2D counterpart, 3D-SIM leverages 3D skeletons, 
% to address the challenges of fine-grained motion and multiple viewpoints
which are inherently viewpoint-agnostic and effective at modeling fine-grained motion.
% enhancing the robustness of the learned representations in scenarios where 2D skeletons (and thus 2D-SIM) struggle. 
Specifically, 3D-SIM performs the auxiliary tasks of feature alignment and feature classification to enrich the visual representations with information from 3D skeletons. %We note that this task is distinct from traditional knowledge distillation (\textcolor{red}{detailed discussion in Section \ref{sec:why_is_kd_different}}). 
Like 2D-SIM, 3D-SIM is only applied during the training phase and removed during inference. An illustration of 3D-SIM is provided in Figure \ref{fig:3dSIM-detailed}.

\textbf{3D Skeleton Representation.}\quad The first step of 3D-SIM is to compute the 3D skeleton representation for a given video. Given a trained 3D skeleton action recognition model, $\mathcal{G}$, 3D-SIM enforces the RGB representations learned by the video transformer to align with 
the skeleton representations from $\mathcal{G}$. In practice, any 3D skeleton-based model can be chosen for $\mathcal{G}$, as long as the temporal and joint dimensions are preserved throughout the model. 
 Given the visual tokens, $\mathbf{z}_l$, of the video transformer layer preceding 3D-SIM, and the corresponding set of 3D skeleton joints:
\begin{equation}
    \mathcal{S}_{3D} = \{(t, j, x, y, x)\} : 1 \leq t \leq T, 1 \leq j \leq J,
\end{equation}
we first obtain 3D skeleton representations as $Y_{3D} = \mathcal{G}(\mathcal{S}_{3D})$, where the shape of $Y_{3D}$ is $T_{s} \times J \times d_{s}$. 

\textbf{Auxiliary Tasks of 3D-SIM.}\quad  Since the goal of 3D-SIM is to learn skeleton representations from visual representations, it employs two hallucination based auxiliary tasks. The first is RGB-skeleton feature alignment, and the second is classifying the hallucinated skeleton features into action classes in the training distribution.
For skeleton feature alignment, we propose two levels of alignment: (1) global, and (2) local. Depending on the level of alignment, we update the skeleton representation $Y_{3D}$ as follows
% \begin{equation}
%     Y_{3D} = 
%     \begin{cases}
%         \frac{1}{T_{s}} \frac{1}{J} \sum_{t=1}^{T_{s}} \sum_{j=1}^{J} Y^{t,j}_{3D} & \textrm{if} \text{ global}\\
%         \frac{1}{J} \sum_{j=1}^{J} Y^{j}_{3D} & \textrm{if} \text{ local}
%     \end{cases}
%     \label{}
% \end{equation}
\begin{equation}
    Y_{3D} = 
    \begin{cases}
        \frac{1}{T_{s}} \frac{1}{J} \sum_{t=1}^{T_{s}} \sum_{j=1}^{J} Y^{t,j}_{3D} & \textrm{if} \text{ global}\\
        \frac{1}{J} \sum_{j=1}^{J} Y^{j}_{3D} & \textrm{if} \text{ local}
    \end{cases}
    % \label{}
\end{equation}

In 3D-SIM, an intermediate visual representation $\Tilde{\mathbf{z}}_l$ is computed for the desired alignment level as
\begin{equation}
    \mathbf{\Tilde{z}}_l = 
    \begin{cases}
        \frac{1}{T_{v}} \frac{1}{S} \sum_{t=1}^{T_{v}} \sum_{h=1}^{S} \mathbf{z}_{t,s} & \textrm{if} \text{ global}\\
        \frac{1}{S} \sum_{h=1}^{S} \mathbf{z}_{s} & \textrm{if} \text{ local}
    \end{cases}
    % \label{}
\end{equation}
For brevity, we ignore the class token in the above equation.
For clarity, we present the shapes of $Y_{3D}$ and $\mathbf{\Tilde{z}}_l$ obtained at different alignment levels in Table~\ref{table:alignment_shapes}. Due to differences in temporal sampling between the visual and skeleton models, $T_{s}$ and $T_{v}$ may not be equal. In the case where $T_{s} > T_{v}$, we uniformly sample $T_{v}$ frames from $Y_{3D}$. In the case where $T_{s} < T_{v}$, we repeat the last frame.

The final step is to project the visual representations $\Tilde{\mathbf{z}}_l$ to the embedding space of the 3D skeleton representation. A parameterized module $f_{3D}()$ can be used to perform the projection:
\begin{equation}
    \hat{Y}_{3D} = f_{3D}(\Tilde{\mathbf{z}}_l)
\end{equation}

Similar to 2D-SIM, this parameterized module $f_{3D}()$ can be implemented using a fully-connected layer that projects $\Tilde{\mathbf{z}}_l$ from $d_{v} \rightarrow d_s$ embedding space. Thus, the shape of $\hat{Y}_{3D}$ is same as $Y_{3D}$ based on the alignment level (see Table~\ref{table:alignment_shapes}). 
Then, the alignment loss of 3D-SIM is computed using Mean Squared Error (MSE) as
\begin{equation}
    \mathcal{L}_{3D}^\text{align} = \frac{1}{\lambda}\sum_{i=1}^{\lambda} (Y_{3D} - \hat{Y}_{3D})^2
\end{equation}
where $\lambda=1$ for global alignment and $\lambda=T$ for local alignment in the common visual skeleton semantic space.
Finally, 3D-SIM performs another auxiliary task of classifying the predicted skeleton features into actions. $\hat{Y}_{3D}$ is processed by a classification head implemented using a FC layer to obtain logits. Then, the cross-entropy loss $\mathcal{L}_{3D}^\text{cls}$ between the logits and ground truth action label of $V$ is computed. This classification auxiliary task improves the discriminability of the predicted skeleton features, thus inducing discriminative 3D skeleton features into the RGB cue. Thus, the total 3D-SIM loss is obtained as
\begin{equation}
    \mathcal{L}_{3D} = \mathcal{L}_{3D}^\text{align} + \mathcal{L}_{3D}^\text{cls}
\end{equation}

\begin{table}[h!]
\centering
\caption{Shape of \( Y_{3D} \) and \( \Tilde{\mathbf{z}}_l \) based on alignment level.}
\begin{tabular}{c|c|c}
\hline
\textbf{Alignment Level} & \textbf{Shape of } \( Y_{3D} \) & \textbf{Shape of } \( \Tilde{\mathbf{z}}_l \) \\
\hline
Global & \( d_{s} \) & \( d_{v} \) \\
Local & \( T_{s} \times d_{s} \) & \( T_{v} \times d_{v} \) \\
%Joint & \( T_{ske} \times J \times d_{ske} \) & \( T_{v} \times J \times d_{vis} \) \\
\hline
\end{tabular}
\label{table:alignment_shapes}
\end{table}
\vspace{-0.35cm}

\subsection{Pose Induced Video Transformer ({\Large$\pi$}-ViT)}

Finally, we integrate 2D-SIM and 3D-SIM together into the video transformer architecture to obtain {\Large$\pi$}-ViT. {\Large$\pi$}-ViT learns unified RGB and pose representations that are view-invariant and can discriminate fine-grained and similar actions, facilitated by the complementary nature of 2D-SIM and 3D-SIM. During the training phase, the modules can be inserted after any layer of the video transformer independently of one another. The total loss that is optimized during training is:
\begin{equation}
    \mathcal{L}_{total} = \mathcal{L}_{v}^{cls} + \mathcal{L}_{2D} + \mathcal{L}_{3D}
\end{equation}
During the inference phase, both modules are removed and only the backbone video transformer architecture is used.

\section{Experimental Results}

\textbf{Datasets.}\quad 
% We evaluate our methods on three popular ADL datasets: \textbf{Toyota-Smarthome}~\cite{smarthome} (Smarthome, SH), \textbf{NTU-120}~\cite{ntu120}, and \textbf{NTU-60}~\cite{NTU_RGB+D}. Smarthome consists of 16K videos recorded in a smarthome environment with a total of 31 actions, we follow the cross-subject (CS) and the two cross-view protocols (CV1, CV2), and measure performance using the mean class accuracy (mCA). NTU-120 contains 114K videos with a total of 120 actions, we follow the cross-subject (CS) and cross-setup protocols (CSet). NTU-60 is a subset of NTU-120 and consists of 57K videos with a total of 60 actions, we follow the cross-subject (CS) and cross-view protocols (CV). Both NTU-60 and NTU-120 are recorded in controlled laboratory environments. We utilize the NTU-60 cross-view-subject (CVS) protocols proposed in~\cite{varol21_surreact} for our ablation studies, as it is a more challenging version of NTU-60. More dataset details can be found in the supplementary.
We assess our methods on three popular ADL datasets: Toyota-Smarthome~\cite{smarthome} (Smarthome, SH), NTU120~\cite{ntu120}, and NTU60~\cite{NTU_RGB+D}. Smarthome comprises 16K videos across 31 actions, using cross-subject (CS) and two cross-view protocols (CV1, CV2), measured by mean class accuracy (mCA). NTU120, with 114K videos and 120 actions, follows CS and cross-setup (CSet) protocols. NTU60, a subset of NTU120, includes 57K videos of 60 actions, using CS and cross-view (CV) protocols. NTU60’s challenging cross-view-subject~\cite{varol21_surreact} (CVS) protocols are used for ablation studies. Further details on the datasets are in the supplementary material.

\noindent\textbf{Implementation details.}\quad In all experiments, we use a pretrained TimeSformer~\cite{timesformer} as the backbone video transformer architecture into which we insert our 2D-SIM and 3D-SIM. On Smarthome, we use Kinetics-400~\cite{kinetics} pretraining and on NTU60 and NTU120, we use Something-Something-v2~\cite{something} pretraining. We follow other RGB+Pose approaches~\cite{das2020vpn, vpn++} and use tracks of human crops as input to our models. Otherwise, we follow the same training procedure as \cite{timesformer}. We use Hyperformer~\cite{zhou2022hyperformer} as our backbone 3D skeleton model $\mathcal{G}$ for 3D-SIM. Default hyperparameter settings for both modules are shown in Table \ref{tab:ablation_studies}.

\subsection{Comparison with State-of-the-art}

We compare our {\Large$\pi$}-ViT with state-of-the-art (SoTA) approaches under three categories: (1) Pose Only, (2) RGB + Pose, and (3) RGB Only (at inference). For our RGB + Pose approach, we perform a late fusion between the logits from {\Large$\pi$}-ViT and Hyperformer. This shows that if 3D Poses are available at inference, our method performs competitively against other RGB + Pose approaches, including PoseC3D~\cite{PoseC3D_CVPR22} which holds SoTA on NTU120 and NTU60.
%We include a RGB + skeleton approach for completeness, but note that our primary contribution is in the RGB only category.

%%%%%%%%%%%%
% SOTA Table Smarthome
%%%%%%%%%%%%
\begin{table}[t]
  \centering
\caption{\textbf{Comparison with SoTA on Toyota-Smarthome dataset.} We report the mean class accuracy on cross-subject (CS) and cross-view (CV$_1$, CV$_2$) protocols. $\circ$ indicates that the modality has been used only in training. Bold text indicates best performance, underline indicates second best performance. $\dagger$ indicates results produced by the authors.}{
\resizebox{0.98\linewidth}{!}{
\begin{tabular}{lccccc}
\hline

\multirow{2}{*}{\textbf{Methods}} & \multicolumn{2}{c}{\textbf{Modality}}  & \multirow{2}{*}{\textbf{CS}} & \multirow{2}{*}{$\textbf{CV}_\mathbf{1}$} & \multirow{2}{*}{$\textbf{CV}_\mathbf{2}$}  \\
 & \textbf{Pose} & \textbf{RGB}  &  \\
 \hline

    %%%% Pose Only methods
    \multicolumn{6}{c}{\cellcolor{gray!20}\textit{Pose Only}} \\ 
    2s-AGCN~\cite{2sagcn2019cvpr}&\checkmark& \xmark &  60.9 & 21.6 & 32.3 \\
    PoseC3D$^\dagger$~\cite{PoseC3D_CVPR22} &\checkmark& \xmark &  50.6 & 20.0 & 28.2\\
    Hyperformer$^\dagger$~\cite{zhou2022hyperformer} &\checkmark& \xmark & 57.5 & 31.6 & 35.2\\
    \hline
    
    %%%% RGB + Pose methods
    \multicolumn{6}{c}{\cellcolor{gray!20}\textit{RGB + Pose}} \\
    P-I3D~\cite{dasWhereToFocus_pi3d_wacv2019} &\checkmark &\checkmark & 54.2 & 35.1 & 50.3\\
    Separable STA~\cite{smarthome} & \checkmark & \checkmark &  54.2  & 35.2 & 50.3  \\
    VPN~\cite{das2020vpn}  & \checkmark & \checkmark  & 65.2 & \underline{43.8} & 54.1  \\
    VPN++ + 3D Poses~\cite{vpn++} & \checkmark & \checkmark  & \underline{71.0} & - & \underline{58.1}  \\
    PoseC3D~\cite{PoseC3D_CVPR22} & \checkmark & \checkmark  & 53.8 & 21.5 &  33.4 \\
    \rowcolor{LightBlue}
    \textbf{{\Large$\pi$}-ViT + 3D Poses (Ours)} & \checkmark & \checkmark & \textbf{73.1} & \textbf{55.6} & \textbf{65.0} \\
    % \rowcolor{LightBlue}
    % TimeSformer~\cite{timesformer}  & \checkmark & \checkmark & \multirow{2}{*}{Fill} & \multirow{2}{*}{Fill} & \multirow{2}{*}{Fill} \\
    % \rowcolor{LightBlue}
    % \cellcolor{red!20}\hspace{0.2cm}\textbf{+ {\Large$\pi$}-ViT (Ours)}  & \checkmark & \checkmark & & &   \\
    \hline

    %%%% RGB only methods
    \multicolumn{6}{c}{\cellcolor{gray!20}\textit{RGB Only (at inference)}} \\
    %2s-AGCN$^*$~\cite{2sagcn2019cvpr} & \checkmark & $\xmark$ & $\xmark$ & - & - & - \\
    %I3D~\cite{i3d} & $\xmark$ & \checkmark & 53.4  & 45.1 \\
    %I3D+NL~\cite{nonlocal} & $\xmark$ & \checkmark &  53.6 &  43.9 \\
    AssembleNet$++$~\cite{assemblenetplusplus} & \xmark & \checkmark &  63.6 & - & -  \\
    LTN~\cite{di_ltn} & \xmark & \checkmark & 65.9  & - & 54.6  \\
    VPN++~\cite{vpn++} & $\circ$ & \checkmark & 69.0 & - & 54.9  \\
    Video Swin$^\dagger$~\cite{liu2021videoswin} & \xmark & \checkmark &   69.8& 36.6 & 48.6\\
    MotionFormer$^\dagger$~\cite{motionformerNeurIPS21} & \xmark & \checkmark & 65.8& 45.2 & 51.0\\
    \hline
    TimeSformer$^\dagger$~\cite{timesformer} & \xmark & \checkmark & 68.4 & 50.0 & 60.6 \\
    \rowcolor{LightBlue}
    \hspace{0.2cm}\textbf{+ 2D-SIM (Ours)}  & $\circ$ & \checkmark & \underline{72.5} & \underline{54.8} & \underline{62.9} \\
    \rowcolor{LightBlue}
    \hspace{0.2cm}\textbf{+ 3D-SIM (Ours)}  & $\circ$ & \checkmark & 71.4 & 51.2 & 62.3 \\
    \rowcolor{LightBlue}
    \textbf{{\Large$\pi$}-ViT (Ours)}  & $\circ$ & \checkmark & \textbf{72.9} & \textbf{55.2} & \textbf{64.8} \\
    \hline

\end{tabular}}
\label{tab:smarthome_sota}
}
\vspace{-1.6em}
\end{table}

\noindent\textbf{Toyota-Smarthome.}\quad In Table \ref{tab:smarthome_sota}, we present the comparison of {\Large$\pi$}-ViT with the SoTA on Smarthome. {\Large$\pi$}-ViT using only RGB at inference achieves SoTA over approaches in all three categories, notably on approaches using both RGB and 3D Poses at inference. {\Large$\pi$}-ViT also achieves a considerable improvement over the TimeSformer backbone, with relative improvements of \textcolor{codegreen}{+6.6\%}, \textcolor{codegreen}{+10.4\%}, and \textcolor{codegreen}{+6.9\%} on CS, CV1, and CV2 protocols respectively. 
%Evaluating each module individually, we see that 2D-SIM leads to larger improvements compared to 3D-SIM (+6.0\% vs +4.4\% on CS). This is logical since Smarthome consists of many fine-grained appearance actions, which 2D-SIM is designed to handle. Additionally, the 3D skeletons provided in Smarthome are estimated and are thus somewhat noisy, which hurts the performance of 3D-SIM. Despite this, we still observe strong improvements of 3D-SIM over both the TimeSformer backbone (+2.8\% on CV$_2$) and Hyperformer backbone (+77.0\% on CV$_2$). See the supplementary for a more detailed discussion on the effect noisy poses have on 2D-SIM and 3D-SIM.
We also find that 2D-SIM and 3D-SIM learn complementary representations, improving the performance of {\Large$\pi$}-ViT over 2D-SIM and 3D-SIM by +0.6\% and +2.0\% respectively on CS protocol. %This indicates that both modules learn representations that target different actions.

%%%%%%%%%%%%
% SOTA Table NTU120
%%%%%%%%%%%%
\begin{table}[t]
\centering
    \caption{\textbf{Comparison with SoTA on NTU120 dataset.} We report the top-1 accuracy on cross-subject (CS) and cross-setup (CS) protocols. $\circ$ indicates that the modality has been used only in training. Bold text indicates best performance, underline indicates second best performance. $\dagger$ indicates results produced by the authors.}{
    \resizebox{0.98\linewidth}{!}{
    \begin{tabular}{lcccc}
    \hline
    
    \multirow{2}{*}{\textbf{Methods}} & \multicolumn{2}{c}{\textbf{Modality}}  & \multirow{2}{*}{\textbf{CS}} &  \multirow{2}{*}{\textbf{CSet}}  \\
     & \textbf{Pose} & \textbf{RGB}  &  &    \\
    
    \hline

    %%%% Pose only methods
    \multicolumn{5}{c}{\cellcolor{gray!20}\textit{Pose Only}} \\ 
    % MS-G3D~\cite{msg3d} &\checkmark& \xmark & 86.9 & 88.4 \\
    InfoGCN~\cite{Hyung-Gun_InfoGCN_CVPR22} &\checkmark& \xmark & 89.8 & 91.2 \\
    PoseC3D~\cite{PoseC3D_CVPR22} &\checkmark& \xmark & 86.0 & 89.6 \\
    Hyperformer~\cite{zhou2022hyperformer} &\checkmark& \xmark & 86.6 & 88.0 \\
    3Mformer~\cite{3mformer_CVPR23} & \checkmark & \xmark & 92.0 & 93.8 \\
    \hline

    %%%% RGB + Pose methods
    \multicolumn{5}{c}{\cellcolor{gray!20}\textit{RGB + Pose}} \\ 
    % P-I3D~\cite{} &\checkmark &\checkmark & &\\
    % Separable STA~\cite{smarthome} & \checkmark & \checkmark &  &  \\
    VPN~\cite{das2020vpn} & \checkmark & \checkmark & 86.3 & 87.8 \\
    VPN++ + 3D Poses~\cite{vpn++} & \checkmark & \checkmark & 90.7 & 92.5 \\
    % ViewCLR + Pose & \checkmark & \checkmark &  &  \\
    PoseC3D~\cite{PoseC3D_CVPR22} &\checkmark& \checkmark & \textbf{95.3} & \textbf{96.4} \\
    STAR-Transformer~\cite{Ahn2022_STARTransformerAR_WACV23} & \checkmark & \checkmark & 90.3 & 92.7 \\
    3D-Def-Transformer~\cite{Kim_3DDeformableAttnForActionRec_ICCV23} & \checkmark & \checkmark & 90.5 & 91.4 \\
    \rowcolor{LightBlue}
    \textbf{{\Large$\pi$}-ViT + 3D Poses (Ours)} & \checkmark & \checkmark & \underline{95.1} & \underline{96.1} \\
    % \rowcolor{LightBlue}
    % TimeSformer~\cite{timesformer}  & \checkmark & \checkmark & \multirow{2}{*}{Fill} & \multirow{2}{*}{Fill} \\
    % \rowcolor{LightBlue}
    % \cellcolor{red!20}\hspace{0.2cm}\textbf{+ {\Large$\pi$}-ViT (Ours)}  & \checkmark & \checkmark &  &  \\
    \hline

    %%%% RGB Only methods
    \multicolumn{5}{c}{\cellcolor{gray!20}\textit{RGB Only (at inference)}} \\ 
    % 2s-AGCN$^*$~\cite{2sagcn2019cvpr} & \checkmark & \xmark & \xmark & - & - & - \\
    %I3D~\cite{i3d} & \xmark & \checkmark & 53.4  & 45.1 \\
    %I3D+NL~\cite{nonlocal} & \xmark & \checkmark &  53.6 &  43.9 \\
    VPN++~\cite{vpn++} & $\circ$ & \checkmark & 86.7 & 89.3 \\
    ViewCLR~\cite{viewclr} & \xmark & \checkmark & 86.2 & 84.5 \\
    ViewCon~\cite{Ketul_ContrastiveMultiview_WACV23} & \xmark & \checkmark & 85.6 & 87.5 \\
    Video Swin$^\dagger$~\cite{liu2021videoswin} & \xmark & \checkmark & 91.4 & 92.1 \\
    MotionFormer$^\dagger$~\cite{motionformerNeurIPS21} & \xmark & \checkmark & 87.0 & 87.9 \\
    
    \hline
    TimeSformer$^\dagger$~\cite{timesformer} & \xmark & \checkmark & 90.6 & 91.6 \\
    \rowcolor{LightBlue}
    \hspace{0.2cm}\textbf{+ 2D-SIM (Ours)}  & $\circ$ & \checkmark & 90.5 & 91.6\\
    \rowcolor{LightBlue}
    \hspace{0.2cm}\textbf{+ 3D-SIM (Ours)}  & $\circ$ & \checkmark & \underline{91.8} & \underline{92.7} \\
    \rowcolor{LightBlue}
    \textbf{{\Large$\pi$}-ViT (Ours)}  & $\circ$ & \checkmark & \textbf{91.9} & \textbf{92.9}\\

\hline
\end{tabular}}
\label{tab:ntu120_sota}
}
\vspace{-1.6em}
\end{table}

\noindent\textbf{NTU120 and NTU60.}\quad We present the comparison of {\Large$\pi$}-ViT with the SoTA on NTU120 and NTU60 in Table \ref{tab:ntu120_sota} and Table \ref{tab:ntu60_sota} respectively. 
On NTU120, {\Large$\pi$}-ViT achieves SoTA compared to other RGB only approaches on the CS and CSet protocols. %{\Large$\pi$}-ViT achieves better performance on STAR-Transformer (+1.8\%/0.2\% on CS/CSet) and VPN++ (+6.0\%/+4.0\% on CS/CSet). 
When compared to the baseline video transformer, {\Large$\pi$}-ViT achieves notable performance boosts by up to \textcolor{codegreen}{+1.4\%} on CSet protocol. 
On NTU60, {\Large$\pi$}-ViT achieves SoTA when compared to other RGB only approaches on the CS protocol, and competitive performance on the CV protocol (97.9\% vs ViewCon's 98.0\%).
% We note that ViewCon is specialized for performance on CV, and {\Large$\pi$}-ViT outperforms ViewCon on CS by +2.6\%.
%Comparing to STAR-Transformer, another video transformer designed for ADL, {\Large$\pi$}-ViT with RGB only achieves a performance increase of +2.2\% and +1.5\% on CS and CV. 
%If RGB + 3D Poses are used, the performance increase is +4.7\% and +2.6\%. 
Consistent with other datasets, we also observe a performance boost of {\Large$\pi$}-ViT over TimeSformer (up to \textcolor{codegreen}{+1.1\%} on CS) backbone. 
%Comparing to VPN++, 3D-SIM achieves a +0.5\% and +1.8\% improvement on CS and CV respectively, indicating 3D-SIMs more powerful use of 3D skeletons. 

\noindent In contrast to Smarthome dataset, we find that 2D-SIM does not improve the action classification performance on NTU datasets. We attribute this to the lack of fine-grained appearance actions in NTU, as these videos are predominantly captured in controlled laboratory settings. 
Nonetheless, the integration of both 2D-SIM and 3D-SIM within the unified architecture of {\Large$\pi$}-ViT facilitates the model's ability to still acquire complementary representations from 3D-SIM.
We also note that PoseC3D~\cite{PoseC3D_CVPR22} outperforms {\Large$\pi$}-ViT on NTU. However, it exhibits significantly lower accuracy in real-world scenarios, particularly in Smarthome environments. This suggests that PoseC3D's effectiveness is heavily dependent on the high quality of pose data.
%Interestingly we find that despite this observation, 2D-SIM and 3D-SIM still learn complementary representations (+0.1\% on CV). 

\noindent \textbf{Other video transformers.} 
We argue that the action recognition performance of the video transformer architectures are already competitive to the preious SoTA~\cite{vpn++}. This corroborates our hypothesis of utilizing video transformers for ADL analysis. 
Across all datasets, {\Large$\pi$}-ViT consistently outperforms all the representative video transformers~\cite{timesformer, liu2021videoswin, motionformerNeurIPS21} substantiating the effectiveness of {\Large$\pi$}-ViT for understanding ADL. 
%In this paragraph, we discuss the performance of our video transformer backbone compared to other approaches on both NTU-60 and NTU-120. One reason {\Large$\pi$}-ViT achieves its performance boosts is because of the use of a strong RGB backbone. For example, TimeSformer alone yields competitive results to VPN++. This helps demonstrates our claim that the effective use of video transformers for ADL is under-explored, and enforces our motivation.
% This observation enforces our motivation that the effective use of video transformers for ADL is under-explored

\begin{figure*}[th]
    \begin{minipage}{0.29\linewidth}
        \centering
        \includegraphics[width=\linewidth]{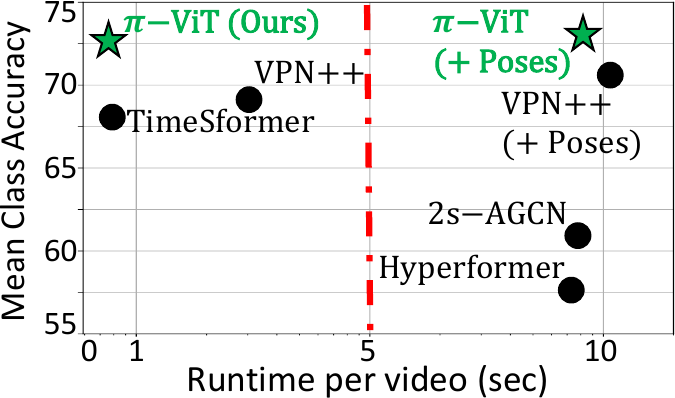}
        \caption{\textbf{Runtime vs accuracy plot} on Toyota-Smarthome. Red dashed line indicates use of 3D poses at inference. 
        % For approaches using pose, we add the runtime of estimating pose to the runtime of the model forward pass.
        }
        \label{fig:runtime-analysis}
    \end{minipage}
    % \hfill
    \hspace{0.5ex}
    \begin{minipage}{0.35\linewidth}
        \centering
        \includegraphics[width=\linewidth]{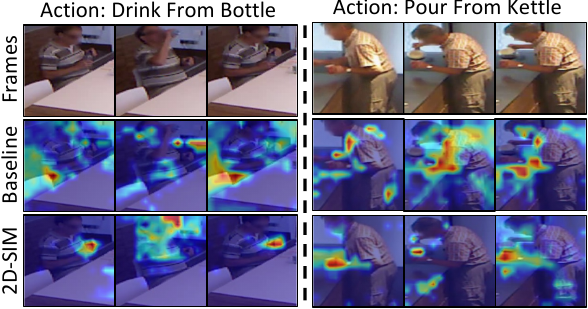}
        \caption{\textbf{GradCAM visualization} of the baseline TimeSformer and our 2D-SIM on two fine-grained appearance actions.
        % from Smarthome. 2D-SIM prioritizes the relevant RGB regions.
        }
        \label{fig:gradcam_visual}
    \end{minipage}
    % \hfill
    \hspace{0.5ex}
    \begin{minipage}{0.35\linewidth}
        \centering
        \includegraphics[width=\linewidth]{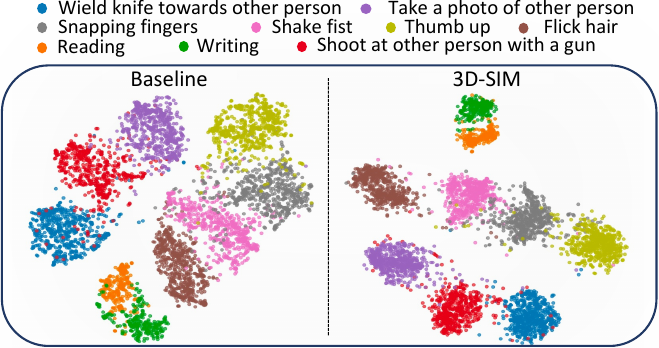}
        \caption{\textbf{TSNE Visualization} of the baseline TimeSformer and our 3D-SIM embeddings on nine visually similar actions.
        % from NTU120. The embeddings of 3D-SIM are more tightly clustered than the baseline.
        }
        \label{fig:tsne-visual}
    \end{minipage}
    % \vspace{-1em}
\end{figure*}

\noindent\textbf{Runtime vs performance.} 
% In Figure \ref{fig:runtime-analysis}, we present a comparison on the runtime and accuracy between our proposed approach, the previous SoTA (VPN++), and pose only approaches on the Toyota-Smarthome dataset. Runtimes encompass the forward pass time as well as the time to extract additional modalities, such as 3D pose. We find that {\Large$\pi$}-ViT outperforms VPN++, both in runtime and performance. Additionally, we observe that {\Large$\pi$}-ViT outperforms approaches utilizing only 3D pose during inference -- at a fraction of the runtime. These accomplishments are due to {\Large$\pi$}-ViT's use of poses during training only, as well as its use of the video transformer backbone, which is computationally cheaper than CNN backbone used in VPN++. In the case that {\Large$\pi$}-ViT has access to 3D poses during inference, it achieves competitive runtimes to pose only approaches while substantially exceeding their performance.
In Fig. \ref{fig:runtime-analysis}, we compare runtime and accuracy of {\Large$\pi$}-ViT with the prior SoTA (VPN++) and pose-only methods on the Toyota-Smarthome dataset. Runtimes include forward pass and modality extraction (e.g., 3D pose) times. We find that {\Large$\pi$}-ViT surpasses VPN++ in both runtime and performance, and outperforms pose-only approaches with significantly shorter runtimes. These accomplishments are due to {\Large$\pi$}-ViT's use of poses during training only. If 3D poses are available to {\Large$\pi$}-ViT during inference, it can match the runtimes of pose only approaches while delivering superior performance.

%%%%%%%%%%% START ABLATION TABLE
\begin{table*}[t]
\centering
\caption{\textbf{Ablation studies.} We ablate the design choices of 2D-SIM and 3D-SIM on the Smarthome cross-subject (SH CS) and NTU60 CVS1 (NTU CVS1) protocols. For each experiment, we highlight the default design choice in gray.}
\label{tab:ablation_studies}

\begin{minipage}{.32\linewidth}
    \centering
    \subcaption{\textbf{Choice of parameterized module.} A simple fully connected layer is sufficient.}
    \resizebox{\linewidth}{!}{
    \begin{tabular}{c|c|c|c|c} \hline 
    \multirow{2}{*}{\textbf{Dataset}} & \multirow{2}{*}{\textbf{Module}} & \multicolumn{3}{c}{\textbf{Choice of $f_\mathbf{2D}$ and $f_\mathbf{3D}$}} \\
    & & \multicolumn{1}{c}{\cellcolor{gray!20}FC} &  \multicolumn{1}{c}{MLP} & \multicolumn{1}{c}{Transformer} \\ \hline
    \multirow{2}{*}{SH CS} & 2D-SIM & 72.5 & 69.9 & 68.4 \\
                 & 3D-SIM & 71.4 & 70.7 & 69.9 \\ \hline
     \multirow{2}{*}{NTU CVS1} & 2D-SIM & 88.0 & 88.74 & 89.1 \\
                 & 3D-SIM & 90.61 & 90.53 & 90.32 \\ \hline
    \end{tabular}}
    \label{subtable:which_parameterized_module}
\end{minipage}%
\hspace{1ex}
\begin{minipage}{.32\linewidth}
    \centering
    \subcaption{\textbf{Comparison of 3D-SIM with traditional distillation.} 3D-SIM's auxiliary alignment is best.}
    \resizebox{\linewidth}{!}{
    \begin{tabular}{l|c|c} \hline 
                \multicolumn{1}{c|}{\textbf{Approach}} & \textbf{SH CS} & \textbf{NTU CVS1} \\ \hline 
                Baseline (TimeSformer) & 68.4 & 86.5 \\
                \hspace{0.1cm}+ FD with class token & 67.1 & 85.1 \\
                \hspace{0.1cm}+ FD with distillation token & 67.6 & 84.4\\
                \hspace{0.1cm}+ LD with class token & 68.3 & 87.9 \\
                \hspace{0.1cm}+ LD with distillation token & 69.1 & 86.7\\
                \cellcolor{gray!20}\hspace{0.1cm}\textbf{+ 3D-SIM (Ours)} & 71.4 & 90.6 \\ \hline 
    \end{tabular}}
    \label{subtable:distillation_vs_3dsim}
\end{minipage}%
\hspace{1ex}
\begin{minipage}{.32\linewidth}
    \centering
    \subcaption{\textbf{Alignment level and position of 3D-SIM.} Global alignment at position $12$ performs best.}
    \resizebox{\linewidth}{!}{
    \begin{tabular}{p{1.2cm}|c|c|c|c|c|c} \hline
                 \multirow{2}{*}{\textbf{Dataset}} & \multirow{2}{*}{\textbf{Baseline}} & \textbf{Alignment} &  \multicolumn{4}{c}{\textbf{3D-SIM Position}}\\ 
                 &  & \textbf{level} &  \multicolumn{1}{c}{1} & \multicolumn{1}{c}{6} & \multicolumn{1}{c}{\cellcolor{gray!20}12} & \multicolumn{1}{c}{1,6,12} \\ \hline
                 \multirow{3}{\linewidth}{SH CS} &  &  \cellcolor{gray!20}Global & 70.3 & 68.9 & 71.4 & 67.7 \\
                 &  68.44&  Local & 68.8 & 68.3 & 70.6 & 69.3\\
                 &  &  Global+Local &  68.0 & 68.3 & 71.3 & 70.5 \\
                 \hline
                 \multirow{3}{\linewidth}{NTU CVS1} &  & \cellcolor{gray!20}Global & 89.0 &  89.0 &  90.6 & 89.7 \\
                 &  86.52&  Local &  88.8 &  89.1 &  90.3 & 89.3\\
                 &  &  Global+Local & 88.5 & 89.3 & 90.1 & 90.6 \\
                 \hline
    \end{tabular}}
    \label{subtable:3dsim_level_position}
\end{minipage}

\vspace{1ex} % Space between the two rows of tables
% Second row of tables
\begin{minipage}{.32\linewidth}
    \centering
    \subcaption{\textbf{Token-skeleton map variants.} The proposed joint-specific token-skeleton map performs best.}
    \resizebox{\linewidth}{!}{
    \begin{tabular}{c|c|c} \hline 
                 \textbf{Variant} &  \textbf{SH CS} & \textbf{NTU CVS1} \\ \hline 
                 \cellcolor{gray!20}Token-Skeleton Map &  72.5 & 88.0 \\ 
                 Flat Variant &  71.1 & 87.8 \\
                 Depth Variant & 71.3 & 87.8 \\ \hline
    \end{tabular}}
    \label{subtable:token-skeleton-map-variants}
\end{minipage}%
\hspace{1ex}
\begin{minipage}{.32\linewidth}
    \centering
    \subcaption{\textbf{Classification task of 3D-SIM.} The classification task yields consistently better performance.}
    \resizebox{\linewidth}{!}{
    \begin{tabular}{c|c|P{1.2cm}|P{1.2cm}} \hline
                 \multirow{2}{*}{\textbf{Dataset}} & \multirow{2}{*}{\textbf{Alignment level}} &  \multicolumn{2}{c}{\textbf{Classification task}}\\ 
                 &  &  \multicolumn{1}{c}{\cellcolor{gray!20}\checkmark} & \multicolumn{1}{c}{\xmark} \\ \hline
                 \multirow{2}{*}{SH CS} & Global & 71.4 & 70.6\\
                 & Local & 70.6 & 70.4 \\
                 \hline
                 \multirow{2}{*}{NTU CVS1}  & Global & 90.3 & 88.6\\
                 & Local & 90.6 & 89.1\\
                 \hline
    \end{tabular}}
    \label{subtable:3dsim_classifier_ablation}
\end{minipage}%
\hspace{1ex}
\begin{minipage}{.32\linewidth}
    \centering
    \subcaption{\textbf{Position of 2D-SIM.} 2D-SIM performs best when placed near the beginning of the model.}
    \resizebox{\linewidth}{!}{
    \begin{tabular}{c|c|c|c|c|c} \hline 
                 \multirow{2}{*}{\textbf{Dataset}} &  \multicolumn{5}{c}{\textbf{2D-SIM Position}}\\
                 &  \multicolumn{1}{c}{\cellcolor{gray!20}1} &  \multicolumn{1}{c}{6} &  \multicolumn{1}{c}{12} &  \multicolumn{1}{c}{1,6} & \multicolumn{1}{c}{1,12} \\ \hline 
                 NTU CVS1 & 88.0 & 87.9 & 87.5 & 87.8 & 87.7 \\
                 SH CS & 72.5 & 70.9 & 69.9 & 70.7 & 70.2  \\ \hline
    \end{tabular}}
    \label{subtable:2dsim_position}
\end{minipage}
\vspace{-1em}
\end{table*}
%%%%%%%%%%%%%%%%%%% END ABLATION TABLE

%%%%%%%%%%%%
% SOTA Table NTU60
%%%%%%%%%%%%
\begin{table}[]
  \centering
\caption{\textbf{Comparison with SoTA on NTU60 dataset.} We report the top-1 accuracy on cross-subject (CS) and cross-view (CV) protocols. $\circ$ indicates that the modality has been used only in training. Bold text indicates best performance, underline indicates second best performance. $\dagger$ indicates results produced by the authors.}{
\resizebox{0.98\linewidth}{!}{
\begin{tabular}{lcccc}
\hline

\multirow{2}{*}{\textbf{Methods}} & \multicolumn{2}{c}{\textbf{Modality}}  & \multirow{2}{*}{\textbf{CS}} &  \multirow{2}{*}{\textbf{CV}}  \\
 & \textbf{Pose} & \textbf{RGB}  &  &    \\

\hline

    %%%% Pose only methods
    \multicolumn{5}{c}{\cellcolor{gray!20}\textit{Pose Only}} \\ 
    % MS-G3D~\cite{msg3d} &\checkmark& \xmark & 91.5 & 96.2 \\
    InfoGCN~\cite{Hyung-Gun_InfoGCN_CVPR22} &\checkmark& \xmark & 93.0 & 97.1 \\
    PoseC3D~\cite{PoseC3D_CVPR22} &\checkmark& \xmark & 93.7 & 96.6 \\
    Hyperformer~\cite{zhou2022hyperformer} &\checkmark& \xmark & 90.7 & 95.1 \\
    3Mformer~\cite{3mformer_CVPR23} & \checkmark & \xmark & 94.8 & 98.7 \\
    \hline

    %%%% RGB + Pose methods
    \multicolumn{5}{c}{\cellcolor{gray!20}\textit{RGB + Pose}} \\ 
    % P-I3D~\cite{} &\checkmark &\checkmark & &\\
    Separable STA~\cite{smarthome} & \checkmark & \checkmark & 92.2 & 94.6 \\
    VPN~\cite{das2020vpn} & \checkmark & \checkmark & 95.5 & 98.0 \\
    VPN++ + 3D Poses~\cite{vpn++} & \checkmark & \checkmark & \underline{96.6} & \underline{99.1} \\
    % ViewCLR + Pose & \checkmark & \checkmark &  &  \\
    PoseC3D~\cite{PoseC3D_CVPR22} &\checkmark& \checkmark & \textbf{97.0} & \textbf{99.6} \\
    STAR-Transformer~\cite{Ahn2022_STARTransformerAR_WACV23} & \checkmark & \checkmark & 92.0 & 96.5 \\
    ViewCon~\cite{Ketul_ContrastiveMultiview_WACV23} & \checkmark & \checkmark & 93.7 & 98.9 \\
    3D-Def-Transformer~\cite{Kim_3DDeformableAttnForActionRec_ICCV23} & \checkmark & \checkmark & 94.3 & 97.9 \\
    \rowcolor{LightBlue}
    \textbf{{\Large$\pi$}-ViT + 3D Poses (Ours)} & \checkmark & \checkmark & 96.3 & 99.0 \\
    % \rowcolor{LightBlue}
    % TimeSformer~\cite{timesformer}  & \checkmark & \checkmark & \multirow{2}{*}{Fill} & \multirow{2}{*}{Fill} \\
    % \rowcolor{LightBlue}
    % \cellcolor{red!20}\hspace{0.2cm}\textbf{+ {\Large$\pi$}-ViT (Ours)}  & \checkmark & \checkmark &  &  \\
    \hline

    %%%% RGB Only methods
    \multicolumn{5}{c}{\cellcolor{gray!20}\textit{RGB Only (at inference)}} \\ 
    % 2s-AGCN$^*$~\cite{2sagcn2019cvpr} & \checkmark & \xmark & \xmark & - & - & - \\
    %I3D~\cite{i3d} & \xmark & \checkmark & 53.4  & 45.1 \\
    %I3D+NL~\cite{nonlocal} & \xmark & \checkmark &  53.6 &  43.9 \\
    Glimpse Clouds~\cite{glimpse} & $\circ$ & \checkmark & 86.6 & 93.0 \\
    VPN++~\cite{vpn++} & $\circ$ & \checkmark & 93.5 & 96.1 \\
    Vyas \textit{et al.}~\cite{Vyas2020MultiviewAR} & \xmark & \checkmark & 82.3 & 86.3 \\
    ViewCLR~\cite{viewclr} & \xmark & \checkmark & 89.7 & 94.1 \\
    Piergiovanni \textit{et al.}~\cite{NPL_2021_CVPR} & \xmark & \checkmark & - & 93.7 \\
    ViewCon~\cite{Ketul_ContrastiveMultiview_WACV23} & \xmark & \checkmark & 91.4 & \textbf{98.0} \\
    % Video Swin$^\dagger$~\cite{liu2021videoswin} & \xmark & \checkmark & 89.35 & 94.02 \\
    % \cellcolor{red!20}MotionFormer$^\dagger$~\cite{motionformerNeurIPS21} & \xmark & \checkmark & 85.72 & Fill \\
    
    % \hline
    % TimeSformer$^\dagger$~\cite{timesformer} & \xmark & \checkmark & 89.31 & 94.31 \\
    % \rowcolor{LightBlue}
    % \hspace{0.2cm}\textbf{+ 2D-{\Large$\pi$}-ViT (Ours)}  & $\circ$ & \checkmark & 89.93 & 94.91 \\
    % \rowcolor{LightBlue}
    % \hspace{0.2cm}\textbf{+ 3D-{\Large$\pi$}-ViT (Ours)}  & $\circ$ & \checkmark & 91.37 & 95.64 \\
    % \rowcolor{LightBlue}
    % \hspace{0.2cm}\textbf{+ {\Large$\pi$}-ViT (Ours)}  & $\circ$ & \checkmark & 91.4 & 95.80 \\
    Video Swin$^\dagger$~\cite{liu2021videoswin} & \xmark & \checkmark & 93.4 & 96.6 \\
    MotionFormer$^\dagger$~\cite{motionformerNeurIPS21} & \xmark & \checkmark & 85.7 & 91.6 \\
    
    \hline
    TimeSformer$^\dagger$~\cite{timesformer} & \xmark & \checkmark & 93.0 & 97.2 \\
    \rowcolor{LightBlue}
    \hspace{0.2cm}\textbf{+ 2D-SIM (Ours)}  & $\circ$ & \checkmark & 93.0 & 97.0 \\
    \rowcolor{LightBlue}
    \hspace{0.2cm}\textbf{+ 3D-SIM (Ours)}  & $\circ$ & \checkmark & \underline{94.0} & 97.8 \\
    \rowcolor{LightBlue}
    \textbf{{\Large$\pi$}-ViT (Ours)}  & $\circ$ & \checkmark & \textbf{94.0} & \underline{97.9} \\

\hline
\end{tabular}}
\label{tab:ntu60_sota}
}
\vspace{-1.6em}
\end{table}

% \begin{figure}
%     \centering
%     \includegraphics[width=0.9\linewidth]{content/gradcam_cropped.pdf}
%     \caption{Gradcam visualization of the Baseline (TimeSformer) and our 2D-SIM.}
%     \label{fig:gradcam_visual}
% \end{figure}

% \begin{figure}
%     \centering
%     \includegraphics[width=\linewidth]{content/gradcam_row_cropped.pdf}
%     \caption{GradCAM visualization of the Baseline (TimeSformer) and our 2D-SIM on two fine-grained appearance actions from Smarthome. 2D-SIM prioritizes the RGB regions that contain the relevant subtle appearance cues.}
%     \label{fig:gradcam_visual}
% \end{figure}

% \begin{figure}
%     \centering
%     \includegraphics[width=0.9\linewidth]{content/heatmap_actions_v4_cropped.pdf}
%     \caption{3D-SIM's performance on visually similar actions. On the x-axis, we present the top-5 NTU-120 actions improved by 3D-SIM. On the y-axis, we present the corresponding four actions that are the most visually similar. For a particular action, the left and right columns indicate number of misclassifications from the baseline and 3D-SIM, respectively. Darker colors are better. The actions are defined by \textbf{A1}: Make Victory Sign, \textbf{A2}: Shoot at Other Person, \textbf{A3}: Reach into Pocket, \textbf{A4}: Writing, \textbf{A5}: Shake Fist. The mean percentage of improvement of 3D-SIM over baseline is shown above the heatmap.}
%     \label{fig:3dsim_heatmap}
% \end{figure}

\subsection{Ablation Study}

In this section, we ablate the design choices of 2D-SIM and 3D-SIM. Ablations are performed on the Smarthome cross-subject (SH CS) and NTU60 CVS1 (NTU CVS1) protocols.

\noindent\textbf{What parameterized module should be used?}\quad %2D-SIM and 3D-SIM make use of parameterized modules, $f_{2D}$ and $f_{3D}$, to perform their respective classification tasks. 
We ablate the choice of parameterized modules, $f_{2D}$ and $f_{3D}$, in Table \ref{subtable:which_parameterized_module}. We evaluate a single FC layer, a 4 layer multi-layer perceptron (MLP), and a 2 layer transformer encoder~\cite{attention} (Transformer). We observe that the performance often improves when a heavier MLP/transformer module is used (2D-SIM on NTU60 CVS1). On the other hand, these modules can also degrade the performance (2D-SIM and 3D-SIM on SH CS). We find that a single fully-connected layer yields strong and consistent performance.

\noindent\textbf{Where should 2D-SIM and 3D-SIM be placed?}\quad In Table \ref{subtable:2dsim_position} and Table \ref{subtable:3dsim_level_position}, we ablate the insertion positions of 2D-SIM and 3D-SIM. 2D-SIM consistently performs best when placed near the initial layers of the transformer, where it has access to low-level, non-contextualized~\cite{attn_rollout} tokens. On the contrary, 3D-SIM performs best when placed near the deeper layers of the transformer, where it has access to more high-level tokens with abstract representations.

\noindent\textbf{How should we construct the token-skeleton map?}\quad We explore two alternative variants of the token-skeleton map and present the results in Table \ref{subtable:token-skeleton-map-variants}. Recall from Section \ref{sec:2dsim} that for a single token, the goal of 2D-SIM is to predict the presence or absence of specific joints within the corresponding RGB patch. In the flat variant of the token-skeleton map, 2D-SIM's task is to predict if the corresponding patch contains \textit{any} joint. In the depth variant, we provide an additional dimension corresponding to depth, and thus 2D-SIM's task is to predict the specific joints contained in the patch, as well as their depth. %Details on the construction of these token-skeleton map variants can be found in the supplementary. 
%This ablation shows that the coarseness of the auxiliary task matters, with the joint-specific map inducing human anatomy provides the best balance. 
This demonstrates that the auxiliary task of token-joint mapping provides relevant human anatomy-based supervision to the RGB cue, while depth does not offer the same level of supervision.

\noindent\textbf{Which alignment level is best for 3D-SIM?}\quad In Table \ref{subtable:3dsim_level_position}, we explore different alignment levels in 3D-SIM. We observe that global alignment consistently yields performance boosts at all positions. In global alignment, the pooling over the temporal dimension mitigates noise and reduces variability, facilitating the auxiliary task of feature alignment. %Additionally, we find that combining global and local (i.e., inserting two 3D-SIM's at the same position, but at different alignment levels) does not lead to improvement. 

\noindent\textbf{Is 3D-SIM's classification task necessary?}\quad %After performing feature prediction, 3D-SIM performs a classification task on top of the predicted features to obtain class predictions. 
In Table \ref{subtable:3dsim_classifier_ablation}, we present the results of 3D-SIM with and without the classification task. We find that the inclusion of this task consistently improves the action classification performance. The additional classification task enforces 3D-SIM to learn discriminative skeleton representation through feature 
alignment, thus inducing more discriminative representation into the RGB representation.

\noindent\textbf{3D-SIM or traditional distillation?}\quad 
In Table \ref{subtable:distillation_vs_3dsim}, we ablate the choice of how to align RGB and 3D skeleton representations. A naive approach is to perform traditional knowledge distillation~\cite{knowledge_distillation_hinton2015, deit} (KD) from a 3D skeleton model to the RGB video transformer. These traditional approaches can be summarized by distilling logits or features into a class token or distillation token. We perform all four combinations of traditional KD and find them ineffective for feature distillation, owing to conflicting gradients between action classification and KD. However, 3D-SIM's feature alignment task in a different embedding space overcomes these KD limitations, thus effectively inducing pose information into video transformers.   
%In traditional KD, alignment is done directly on the RGB feature space. This is contrary to 3D-SIM, which 
%3D-SIM applies a learned transformation on the RGB space and then performs alignment which provides the RGB representations more degrees of freedom, allowing them to converge to a space that is effective for both the primary video transformer task and the 3D-SIM auxiliary task. This is evidenced by the performance gains observed by 3D-SIM compared to traditional KD.

\subsection{Does {\Large$\pi$}-ViT Address ADL Challenges?}

In this section, we provide an analysis of 2D-SIM and 3D-SIM on the challenges of ADL. %Recall that 2D-SIM is designed to address the challenge of fine-grained appearance actions, while 3D-SIM addresses the challenge of fine-grained motion and challenging viewpoints.
In Figure \ref{fig:gradcam_visual}, we compare the Gradient Class Activation Maps (GradCAM)~\cite{grad_cam} of the baseline TimeSformer against 2D-SIM on two fine-grained action videos from Smarthome. We observe that 2D-SIM prioritizes RGB regions containing the relevant human joints that characterize the actions. %Specifically, we see that 2D-SIM prioritizes regions containing the human joints relevant to the action, as well as the regions containing relevant objects. For both actions, we observe that the baseline model does not prioritize the relevant RGB regions.
In Figure \ref{fig:tsne-visual}, we use T-SNE~\cite{tsne} to visualize embeddings of the baseline TimeSformer and 3D-SIM on nine visually similar actions from NTU120. We find that 3D-SIM learns more tightly packed clusters than TimeSformer, highlighting its discriminative power and ability to disambiguate visually similar actions.

% Figure \ref{fig:3dsim_heatmap} presents a heatmap visualization comparing the frequency of misclassifications for visually similar actions with subtle motion between 3D-SIM and the baseline TimeSformer. It distinctly shows that 3D-SIM effectively disambiguates these actions, as indicated by the darker columns.    
% Furthermore, as observed in Tables \ref{tab:smarthome_sota}, \ref{tab:ntu120_sota}, and \ref{tab:ntu60_sota}, 3D-SIM consistently improves performance across all cross-view protocols using only RGB at inference, demonstrating its viewpoint invariance.
%To show that 3D-SIM is effective, we consider the pair-wise misclassification frequencies between different actions in NTU-120. We present this as a heatmap visualization in Figure \ref{fig:3dsim_heatmap}. To select actions for the x-axis, we choose the top-5 actions where 3D-SIM leads to improvement. For the y-axis, we examine the corresponding top-10 actions that the baseline TimeSformer misclassifies, and manually select the 5 actions are the most visually similar in their motion. The intensity of the color indicates the number of misclassifications, and thus darker colors are better. For a given action, the left and right columns indicate the number of misclassifications by the baseline and 3D-SIM, respectively. Regarding viewpoint invariance, we observe in Table \ref{tab:smarthome_sota}, Table \ref{tab:ntu120_sota}, and Table \ref{tab:ntu60_sota} that 3D-SIM yields consistent improvement on all cross-view protocols, demonstrating its invariance to viewpoint changes.

\section{Related Works}
% In recent years, vision transformers~\cite{dosovitskiy2020vit, deit, liu2021swin, mvit1} have overtaken CNNs~\cite{resnet, vgg16, szegedy2016inception} in performance across numerous image-based tasks. Similarly, video transformers~\cite{timesformer, liu2021videoswin, vivit, motionformerNeurIPS21, mvit1, mvit2}  have had a comparable effect on 3DCNNs~\cite{x3d, lin2019tsm, i3d} and two-stream CNNs for video-based tasks~\cite{twostream, twostreamfusion, slow_fast}.
% These video transformers are tailored for analyzing web-based videos~\cite{kinetics, ucf, kuehne2011hmdb, AVA}, consisting of prominent motion patterns and frame-centric actions, and often fall short when dealing with ADL. ADL videos~\cite{smarthome, charades, NTU_RGB+D, ntu120, MSRDailyactivity3D} present challenges that video transformers are not designed to handle, such as visually similar actions and challenging views.
% To address these challenges, previous work~\cite{stgcn, msaagcn, Hyung-Gun_InfoGCN_CVPR22, hachiuma2023unifiedskele} has proposed approaches that utilize only human poses. These approaches are effective on datasets recorded in laboratory settings~\cite{NTU_RGB+D, ntu120, nucla} where motion is the dominant cue in distinguishing actions. However, they struggle with real-world videos~\cite{vpn++, smarthome} where motion and appearance are required. In response, various approaches~\cite{Kim_3DDeformableAttnForActionRec_ICCV23, Ahn2022_STARTransformerAR_WACV23, das2020vpn} have used both RGB and pose modalities to model ADL.

Recent advancements in vision transformers~\cite{dosovitskiy2020vit, deit, liu2021swin, mvit1} have surpassed CNNs~\cite{resnet, vgg16, szegedy2016inception} on image-based tasks. Similarly, video transformers~\cite{timesformer, liu2021videoswin, vivit, mvit1, mvit2} have excelled over 3DCNNs~\cite{x3d, lin2019tsm, i3d} and two-stream CNNs~\cite{twostream, twostreamfusion, slow_fast} in video tasks. These video transformers, optimized for web videos, struggle on ADL videos~\cite{smarthome, charades, NTU_RGB+D, ntu120, MSRDailyactivity3D}, which pose unique challenges. Approaches using human poses~\cite{stgcn, msaagcn, Hyung-Gun_InfoGCN_CVPR22, hachiuma2023unifiedskele} are effective in laboratory settings~\cite{NTU_RGB+D, ntu120, nucla} but limited in real-world videos~\cite{smarthome, charades}. 

Therefore, several approaches combine the RGB and pose modalities~\cite{Kim_3DDeformableAttnForActionRec_ICCV23, Ahn2022_STARTransformerAR_WACV23, das2020vpn} to address the challenges of ADL. Recently, STAR-Transformer~\cite{Ahn2022_STARTransformerAR_WACV23} and 3D deformable transformer~\cite{Kim_3DDeformableAttnForActionRec_ICCV23} have effectively utilized these modalities within video transformers.   
Both STAR-Transformer and 3D deformable transformer introduce specialized spatio-temporal attention mechanisms to enable cross-modal learning in video transformers. These methods capitalize on the efficacy of video transformers discussed above.
% perform cross-modal learning in video transformers through the introduction of specialized spatio-temporal attention mechanisms. These methods capitalizes on the efficacy of video transformers discussed above.

While the above approaches combining RGB and pose are effective, the burden of collecting poses at inference time is high, as specialized sensors or expensive pose estimation is required. Thus, approaches~\cite{vpn++, NPL_2021_CVPR, glimpse} using only the RGB are desirable. Our proposed {\Large$\pi$}-ViT is one such approach that does not require any pose information at inference.
Closest to our work is VPN++~\cite{vpn++}, which integrates 3D pose information into a CNN-based RGB backbone through feature and attention level distillations, and does not require poses at inference. However, our experiments show that these distillations are sub-optimal in video transformers. Moreover, the previous RGB + pose approaches overlook the correspondence between 2D skeleton joints and spatial regions of interest. In contrast, {\Large$\pi$}-ViT induces both 2D and 3D pose information within the RGB cue in video transformers through pose-aware auxiliary tasks.

%is one such approach that is the closest to our own. To remove the requirement of pose at inference, VPN++ uses feature-level and attention-level distillation to infuse 3D pose information into a CNN-based RGB backbone. However, VPN++ uses only 3D poses and does not explicitly address the challenge of fine-grained appearance in ADL. Additionally, VPN++ uses a CNN backbone and does not benefit from the more powerful video transformer.

%\noindent\textbf{Video transformers for ADL.}\quad Despite the success of video transformers over CNNs, few works~\cite{Ahn2022_STARTransformerAR_WACV23, Kim_3DDeformableAttnForActionRec_ICCV23} have utilized them for understanding ADL. One potential reason for this is that since the architecture designs are different, methods that work on CNNs typically do not work out-of-the-box on video transformers. As for current approaches, STAR-Transformer~\cite{Ahn2022_STARTransformerAR_WACV23} proposes alternative spatio-temporal attention mechanisms within the transformer to efficiently represent cross-modal features from RGB and human pose. Kim et al.~\cite{Kim_3DDeformableAttnForActionRec_ICCV23} proposes a 3D deformable transformer leveraging specialized attention mechanisms and a cross-modal learning scheme for RGB and human pose. However, both of these approaches require human poses at inference time, and do not explicitly address the challenges of ADL.

\section{Conclusion}

% In conclusion, we propose {\Large$\pi$}-ViT, the first approach for understanding ADL that leverages both 2D and 3D human poses and only requires RGB for inference.
In conclusion, we propose {\Large$\pi$}-ViT, the first video transformer model leveraging both 2D and 3D human poses for understanding ADL videos. {\Large$\pi$}-ViT consists of two novel plug-in modules, 2D-SIM and 3D-SIM, which are inserted into a video transformer model. Each module performs a distinct auxiliary task that induces human pose knowledge into the RGB representation space of the model, enabling {\Large$\pi$}-ViT to address the specific challenges of ADL. Notably, the modules are only required during training and thus there is no requirement of poses at inference, drastically reducing the computational cost of {\Large$\pi$}-ViT. We show that {\Large$\pi$}-ViT effectively addresses the challenges of ADL and achieves state-of-the-art performance on popular ADL datasets.

\newpage
\section{Acknowledgements}
We thank Vishal Singh for his valuable assistance in implementing and executing methods used in our SoTA comparisons. Additionally, we thank the members of the Charlotte Machine Learning Lab at UNC Charlotte for helpful discussions. This work is partially supported by the National Science Foundation (IIS-2245652). Any opinions, findings, conclusions or recommendations expressed in this material are those of the authors and do not necessarily reflect the views of the funders.

%%%%%%%%%%%%%%%%%%%%%%%%%%%%%%%%%%%%%%%%%%%%%%%%%%%%
%% APPENDIX
%%%%%%%%%%%%%%%%%%%%%%%%%%%%%%%%%%%%%%%%%%%%%%%%%%%%
\section*{\Large Appendix}
\appendix

\section{Datasets and Evaluation Protocols}
We evaluate our methods on three popular Activities of Daily Living (ADL) datasets. 

\noindent \textbf{Toyota-Smarthome}~\cite{smarthome} (Smarthome, SH) provides 16.1K video clips of elderly individuals performing actions in a real-world smarthome setting. The dataset contains 18 subjects, 7 camera views, and 31 action classes. For evaluation, we follow the cross-subject (CS) and cross-view (CV$_1$, CV$_2$) protocols. Due to the unbalanced nature of the dataset, we use the mean class-accuracy (mCA) performance metric. The dataset provides 2D and 3D skeletons containing $13$ keypoints that were extracted using LCRNet \cite{lcrnet_new}, which are used to generate the inputs to our 2D-SIM and 3D-SIM approaches.

\noindent \textbf{NTU120}~\cite{ntu120} provides 114K video clips of subjects performing actions in a controlled laboratory setting. The dataset consists of 106 subjects, 3 camera views, and 120 action classes. We follow the cross-subject (CS) and cross-setup (CSet) protocols for evaluation, and report the top-1 classification accuracy. The dataset provides 2D and 3D skeletons containing $25$ keypoints extracted using Microsoft Kinect v2 sensors, which we use to generate the inputs to our 2D-SIM and 3D-SIM approaches.

\noindent \textbf{NTU60}~\cite{NTU_RGB+D} is a subset of NTU120 that provides 56.8K video clips of subjects performing actions in a controlled laboratory setting. The dataset consists of 40 subjects, 3 camera views, and 60 action classes. For evaluation, we follow the cross-subject (CS) and cross-view (CV) protocols, and report the top-1 accuracy. For our ablations, we follow the cross-view-subject (CVS) protocol, CVS1, as proposed in \cite{varol21_surreact}. In the CVS protocols, the subjects and viewpoints in the training set are distinct from the subjects and viewpoints in the testing set. Specifically, only the $0^\circ$ viewpoint from the NTU60 CS training protocol is used for training, while testing is carried out on the $0^\circ$, $45^\circ$, or $90^\circ$ viewpoints from the NTU60 CS test split, which are referred to as CVS1, CVS2, and CVS3, respectively. We use the CVS protocols because they provide a better represent the cross-view challenge. The dataset provides 2D and 3D skeletons containing $25$ keypoints extracted using Microsoft Kinect v2 sensors, which we use to generate the inputs to our 2D-SIM and 3D-SIM approaches.

\section{Implementation Details (Additional)}
We train all our models on $8$ RTX A5000 or A6000 GPUs.

\noindent\textbf{Our models.}\quad In all experiments, we use a $12$ layer TimeSformer~\cite{timesformer} video transformer backbone and follow a training pipeline similar to \cite{timesformer}. We use Kinetics400 pretraining for Smarthome and SSv2 pretraining for NTU60 and NTU120. For fine-tuning, we train our models for $15$ epochs. The RGB inputs to our models are video frames of size $8\times224\times224$ for Smarthome and a size of $16\times224\times224$ for NTU60 and NTU120. Frames are sampled at a rate of $\frac{1}{32}$ for Smarthome and uniform sampling is used for NTU60 and NTU120. As done in \cite{das2020vpn, vpn++}, we extract $224 \times 224$ human crops from the video before feeding them to our models. This ensures that the video frames input to our model will contain human skeleton joints.

\begin{figure}[h]
    \centering
    \includegraphics[width=0.25\linewidth]{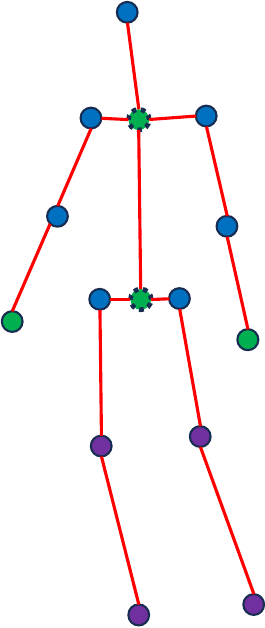}
    \caption{Illustration of the human joint partitions used when training Hyperformer on Smarthome. Color denotes partition, dashed outline indicates interpolated human joint.}
    \label{fig:hypergraph-sh}
\end{figure}

\begin{figure*}[th]
    \centering
    \begin{subfigure}{0.48\linewidth}
        \includegraphics[width=\linewidth]{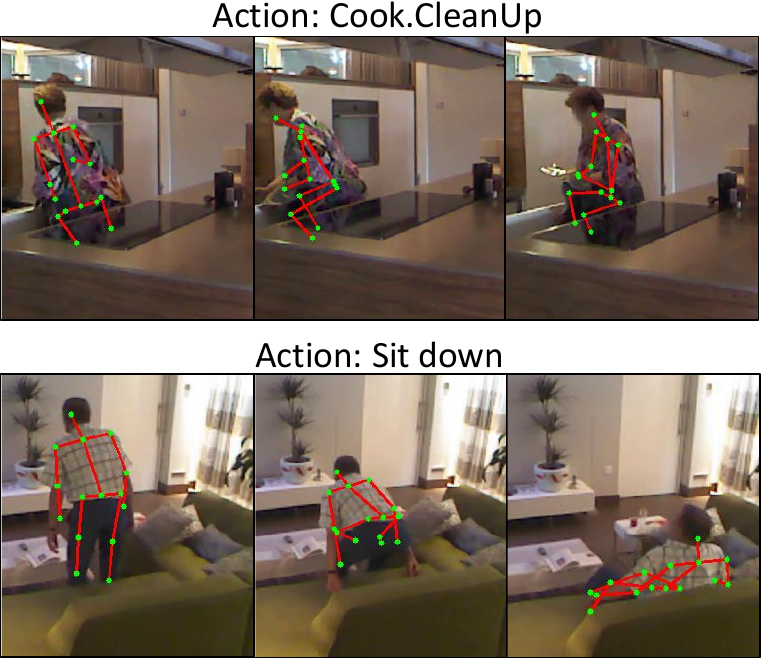}
        \caption{Sample videos and poses from Toyota-Smarthome.}
    \end{subfigure}
    \hfill
    \begin{subfigure}{0.48\linewidth}
        \includegraphics[width=\linewidth]{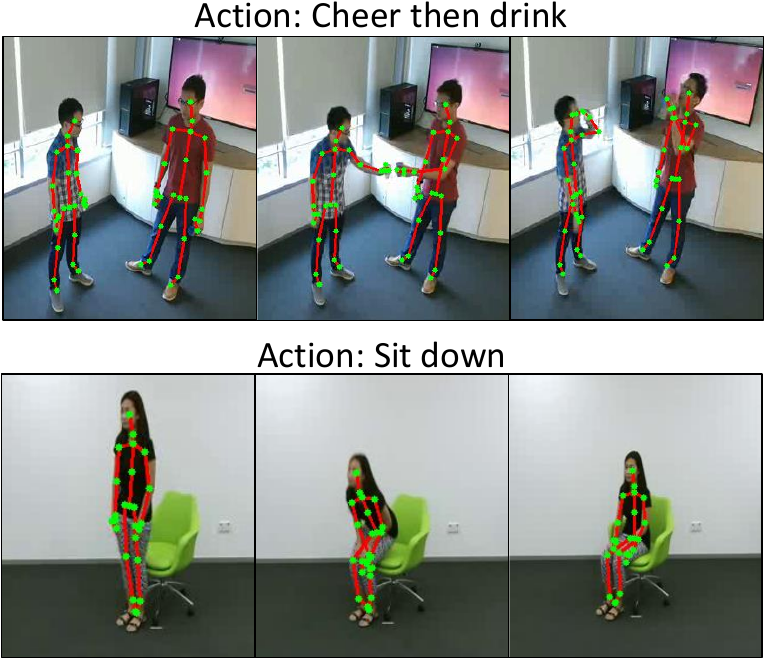}
        \caption{Sample videos and poses from NTU.}
    \end{subfigure}
    \caption{Visualizations of poses from Smarthome (a) and NTU (b).}
    \label{fig:pose-visualizations}
\end{figure*}

\noindent\textbf{Other video transformers.}\quad For results we generate ourselves using other video transformer methods~\cite{liu2021videoswin, motionformerNeurIPS21} (indicated by $\dagger$ in SoTA tables), we follow the default configurations suggested by each method. For a fair comparison with our models, we also utilize Kinetics400 pretraining for Smarthome and SSv2 pretrainining for NTU60 and NTU120.

\noindent\textbf{3D skeleton model.}\quad As mentioned in the main paper, we use Hyperformer~\cite{zhou2022hyperformer} as the pretrained 3D skeleton model in 3D-SIM. 
%Hyperformer's codebase provides pretrained models for the protocols of NTU60 and NTU120, but not for Smarthome. 
For Smarthome, we train Hyperformer using the human joint partition shown in Figure \ref{fig:hypergraph-sh}, otherwise we follow the same training configuration proposed in \cite{zhou2022hyperformer}. Note that we interpolate the base-of-spine and top-of-spine keypoints. This allows the origin of the joints to be centered at the spine, a required pre-processing step in Hyperformer.

\begin{figure}[h]
    \centering
    \begin{subfigure}{\linewidth}
        \includegraphics[width=\textwidth]{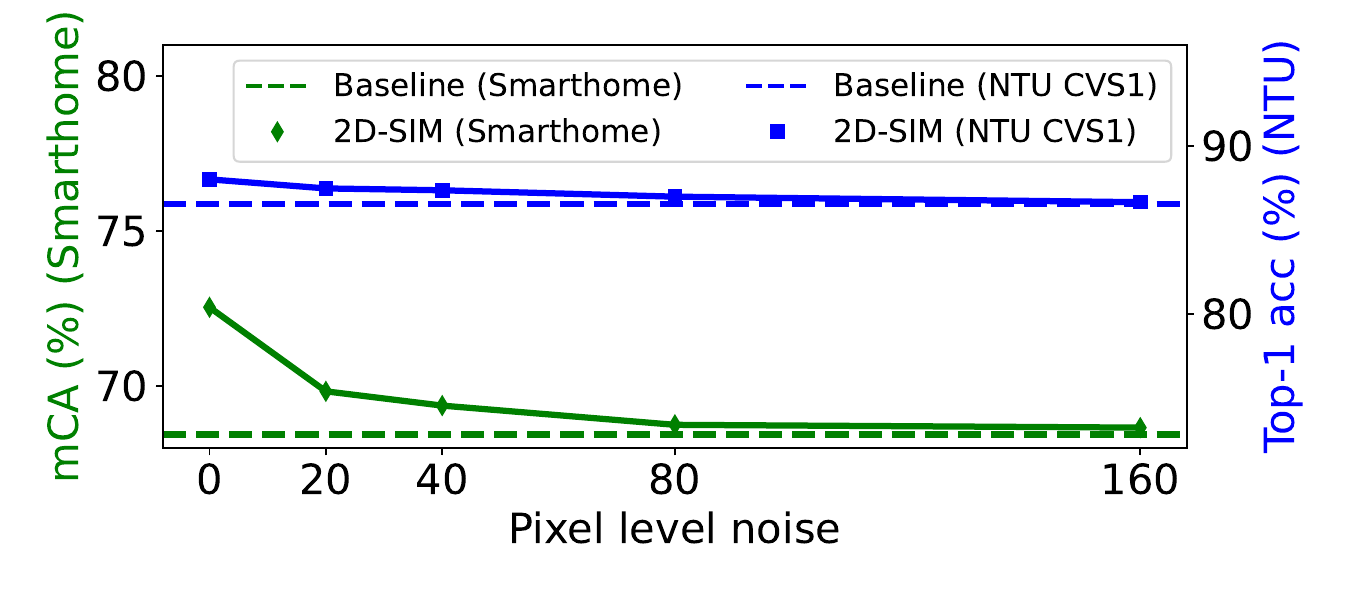}
        \caption{2D-SIM}
        \label{fig:noisy-pose-2dsim}
    \end{subfigure}
    \hfill
    \begin{subfigure}{\linewidth}
        \includegraphics[width=\textwidth]{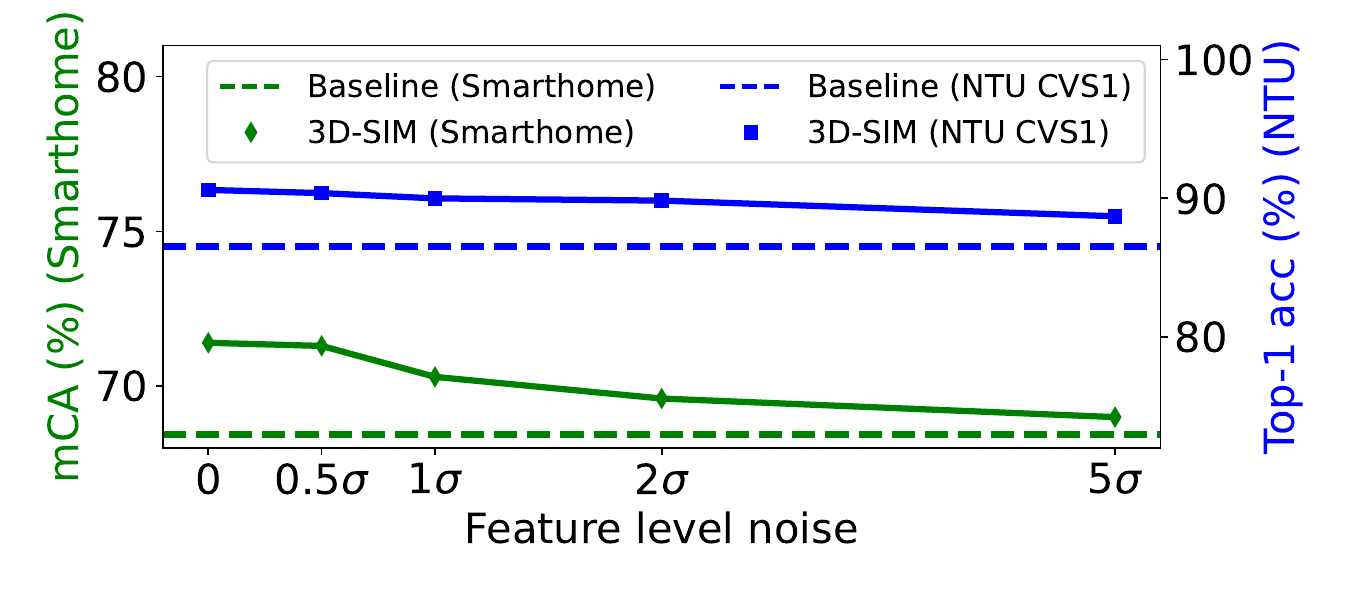}
        \caption{3D-SIM}
        \label{fig:noisy-pose-3dsim}
    \end{subfigure}
    \caption{Effects of noisy poses on 2D-SIM and 3D-SIM on Smarthome CS and NTU CVS1 protocols.}
    \label{fig:noisy-pose-effect}
\end{figure}

\section{The Effects of Noisy Poses}
% In Figure \ref{fig:pose-visualizations} we show visualizations of 2D human poses from Smarthome and NTU. In the real-world, capturing human poses is not an easy task and Smarthome exeplifies this. In Smarthome there can be occlusions and odd viewing angles, which we observe impacts the quality of the estimated poses. In NTU the poses do not have these problems. The reasons are that NTU poses are collected using a specialized sensor, and that NTU actions are highly choreographed, with the subject always completely visible to the camera with no occlusions.
Figure \ref{fig:pose-visualizations} highlights the challenges encountered in real-world human pose estimation, particularly evident in the Smarthome dataset. Issues such as occlusions and unusual camera angles frequently degrade the accuracy of pose estimations in such settings. It is worthwhile to mention that datasets like NTU tend to exhibit fewer of these complications due to their controlled collection environments and use of specialized sensors for collecting poses, both of which are impractical in the real world. 

These observations exemplify the need to design algorithms that are robust to noisy poses. In Figure \ref{fig:noisy-pose-effect}, we introduce varying levels of noise into 2D-SIM and 3D-SIM to evaluate their effectiveness in the presence of noisy poses. For 2D-SIM, we directly introduce pixel-level noise into the human skeleton joints. For each joint coordinate, we randomly sample two values between $0$ and the designated noise level and add it to the joints $x$ and $y$ coordinates. For 3D-SIM, we add noise to the 3D skeleton features used as input to the model. The levels of noise chosen are based on the standard deviations of the features, and then, similarly to 2D-SIM, we randomly sample values between $0$ and the designated noise level and add it to the feature vector to generate noisy 3D skeleton features.

In Figure \ref{fig:noisy-pose-2dsim}, we observe that 2D-SIM is sensitive to very noisy poses, causing the performance to match the baseline. This makes sense as with wildly inaccurate poses, the extra supervision provided by 2D-SIM will be wasted on non-salient RGB regions. At low to medium levels of noise ($20$, $40$), which are more likely to apply in real-world pose estimation, 2D-SIM still provides improvements over the baseline. Figure \ref{fig:noisy-pose-3dsim} shows the effect of noisy 3D skeleton features on 3D-SIM. We see that 3D-SIM is more robust to high levels of noise, consistently outperforming the baseline across all noise levels. This can be attributed to the inherent robustness of 3D skeleton models to noisy poses, as shown in previous research~\cite{PoseC3D_CVPR22}.

\begin{table}[]
    \centering
    \caption{Top-5 classes improved by 2D-SIM and 3D-SIM on Toyota-Smarthome CS and NTU120 CS.}
    \label{tab:absolute_improvements}
    \begin{minipage}{.51\linewidth}
        \centering
        \subcaption{2D-SIM}
        \resizebox{0.999\linewidth}{!}{
            \begin{tabular}{cc}
                \hline
                 \multirow{2}{*}{\textbf{Action name}} &  \textbf{Improvement} \\
                  &  \textbf{over baseline} \\
                 \hline
                 \multicolumn{2}{c}{\cellcolor{gray!20}\textit{Toyota-Smarthome}} \\
                Drink.FromGlass & +33.3\% \\
                Use Tablet & +13.3\% \\
                Drink.Frombottle & +11.4\% \\
                WatchTV & +10.0\% \\
                MakeCoffee.PourGrains & +9.5\% \\
                 \hline
                 \multicolumn{2}{c}{\cellcolor{gray!20}\textit{NTU120}} \\
                Cut Paper w/ Scissors & +4.0\% \\
                Reading & +2.9\% \\
                Thumb down & +2.4\% \\
                Shoot at basket & +2.3\% \\
                Clapping & +2.2\% \\
                 \hline
            \end{tabular}
            }
            \label{tab:abs-improvements-2dsim}
    \end{minipage}%
    \hfill
    \begin{minipage}{.49\linewidth}
    \centering
    \subcaption{3D-SIM}
    \resizebox{0.97\linewidth}{!}{
        \begin{tabular}{cc}
            \hline
             \multirow{2}{*}{\textbf{Action name}} &  \textbf{Improvement} \\
              &  \textbf{over baseline} \\
             \hline
             \multicolumn{2}{c}{\cellcolor{gray!20}\textit{Toyota-Smarthome}} \\
            Eat Snack & +13.7\% \\
            Maketea.Boilwater & +12.5\% \\
            Cook.Usestove & +11.1\% \\
            Pour.Frombottle & +10.6\% \\
            WatchTv & +10.0\% \\
             \hline
             \multicolumn{2}{c}{\cellcolor{gray!20}\textit{NTU120}} \\
            Rub hands together & +6.9\% \\
            Make victory sign & +5.0\% \\
            Wield knife at person & +4.5\%\\
            Yawn & +4.4\% \\
            Play magic cube & +3.7\% \\
             \hline
        \end{tabular}
        }
        \label{tab:abs-improvements-3dsim}
\end{minipage}%
\end{table}

\begin{table}
    \centering
    \caption{Top-5 class-pairs improved by 2D-SIM over the Baseline (TimeSformer) on Toyota-Smarthome CS and NTU120 CS.}
    \resizebox{0.98\linewidth}{!}{
    \begin{tabular}{ccc}
        \hline
         \multirow{2}{*}{\textbf{Action 1}} &  \multirow{2}{*}{\textbf{Action 2}} & \textbf{2D-SIM Improvement} \\
         & & \textbf{over baseline} \\
         \hline
         \multicolumn{3}{c}{\cellcolor{gray!20}\textit{Toyota-Smarthome}} \\
        Takepills & UseTelephone & +61.5\% \\
        Pour.Frombottle & Pour.Fromcan & +60.0\% \\
        Drink.Frombottle & Drink.Fromglass & +57.1\% \\
        WatchTV & ReadBook & +40.6\% \\
        Drink.FromCan & WatchTV & +29.4\% \\
         \hline
         \multicolumn{3}{c}{\cellcolor{gray!20}\textit{NTU120}} \\
         Toss coin & Make ok sign & +38.9\% \\
         Reading & Writing & +23.5\% \\
         Make victory sign & Make ok sign & +14.3\% \\
         Yawn & Blow nose & +13.6\% \\
         Cut Paper w/ Scissors & Staple book & +12.7\% \\
         \hline
    \end{tabular}
    }
    \label{tab:failure-cases-2dsim}
\end{table}

\begin{table}
    \centering
    \caption{Top-5 class-pairs improved by 3D-SIM over the Baseline (TimeSformer) on Toyota-Smarthome CS and NTU120 CS.}
    \resizebox{0.98\linewidth}{!}{
    \begin{tabular}{ccc}
        \hline
         \multirow{2}{*}{\textbf{Action 1}} &  \multirow{2}{*}{\textbf{Action 2}} & \textbf{3D-SIM Improvement} \\
         & & \textbf{over baseline} \\
         \hline
         \multicolumn{3}{c}{\cellcolor{gray!20}\textit{Toyota-Smarthome}} \\
         WatchTV & UseTelephone & +42.85\%\\
         WatchTV & ReadBook & +33.33\% \\
         Cook.Cleanup & Walk & +29.41\% \\
         Cook.Cleandishes & Cook.Cleanup & +15.15\% \\
         Enter & Leave & +7.10\% \\
         \hline
         \multicolumn{3}{c}{\cellcolor{gray!20}\textit{NTU120}} \\
         Yawn& Flick hair& +54.55\% \\
         Rub hands together& Clapping& +32.43\% \\
         Yawn&  Blow nose&  +25.76\% \\
         Make victory sign&  Make okay sign& +25.51\% \\
         Cut paper& Staple book& +13.64\% \\
         \hline
    \end{tabular}
    }
    \label{tab:failure-cases-3dsim}
\end{table}

\section{Improvement Cases} In Table \ref{tab:absolute_improvements}, the top-5 action classes demonstrating significant performance enhancements via 2D-SIM and 3D-SIM over the baseline in the Toyota-Smarthome CS and NTU120 CS protocols are presented. Notably, the largest improvements of 2D-SIM are observed in actions with fine-grained appearance details, such as \textit{Drink from glass} and \textit{Use Tablet}. This indicates the effectiveness of 2D-SIM on actions where modeling fine-grained appearance is necessary. For 3D-SIM, the largest improvements come from actions with fine-grained motion, e.g., \textit{rub hands together}.

In Table \ref{tab:failure-cases-2dsim} and Table \ref{tab:failure-cases-3dsim}, we present the the top-5 class-pairs that are improved by 2D-SIM and 3D-SIM on the Toyota-Smarthome CS and NTU120 CS protocols. The metric displayed is the raw number of predictions, i.e., the number Action 1 samples that were misclassified as Action 2. We observe that the baseline often confuses actions with similar appearance, such as confusing \textit{Takepills} with \textit{UseTelephone} or \textit{Drink.Frombottle} with \textit{Drink.Fromglass}. We also observe that 2D-SIM can improve performance in such cases, owing to the additional supervision applied to the salient RGB regions. We also observe that the baseline confuses actions with similar motion, such as \textit{Yawn} vs \textit{Blow nose}, and show that 3D-SIM improves the performance in these cases.

\begin{table}[h]
    \centering
    \caption{Comparison of our methods to the 3D skeleton model used in 3D-SIM (Hyperformer) and the baseline TimeSformer.}
    \label{tab:all-comparison}
    \resizebox{0.98\linewidth}{!}{
    \begin{tabular}{l|ccc|cc|cc}
    \hline
    \multirow{2}{*}{\textbf{Method}}& \multicolumn{3}{c|}{\cellcolor{gray!20}Toyota-Smarthome}& \multicolumn{2}{c|}{\cellcolor{gray!20}NTU60}& \multicolumn{2}{c}{\cellcolor{gray!20}NTU120}\\
     &  \textbf{CS}&  \textbf{CV}$_\mathbf{1}$&  \textbf{CV}$\mathbf{_2}$&  \textbf{CS} &  \textbf{CV} &  \textbf{CS} & \textbf{CSet}\\
     \hline
     {\color[HTML]{9B9B9B}Hyperformer~\cite{zhou2022hyperformer}} & {\color[HTML]{9B9B9B}57.5} & {\color[HTML]{9B9B9B}31.6} & {\color[HTML]{9B9B9B}35.2} & {\color[HTML]{9B9B9B}90.7} & {\color[HTML]{9B9B9B}95.1} & {\color[HTML]{9B9B9B}86.6} & {\color[HTML]{9B9B9B}88.0} \\
    \textbf{{\Large$\pi$}-ViT + 3D Poses}&  73.1&  55.6&  65.0&  96.3 &  99.0 & 95.1& 96.1\\
    \hline
     {\color[HTML]{9B9B9B}TimeSformer~\cite{timesformer}} &  {\color[HTML]{9B9B9B}68.4}&  {\color[HTML]{9B9B9B}50.0}&  {\color[HTML]{9B9B9B}60.6}&  {\color[HTML]{9B9B9B}93.0} &  {\color[HTML]{9B9B9B}97.2} & {\color[HTML]{9B9B9B}90.6}& {\color[HTML]{9B9B9B}91.6} \\
     \hspace{0.2cm}\textbf{+ 2D-SIM} &  72.5&  54.8&  62.9&  93.0 &  97.0 & 90.5& 91.6\\
     \hspace{0.2cm}\textbf{+ 3D-SIM} &  71.4&  51.2&  62.3&  94.0 &  97.8 & 91.8& 92.7\\
     \textbf{{\Large$\pi$}-ViT} & 72.9&  55.2&  64.8&  94.0 &  97.9 & 91.9& 92.9\\
     \hline
    \end{tabular}
    }
\end{table}

\section{Comparison with baseline and 3D skeleton model}
In Table \ref{tab:all-comparison}, we compare our methods with the baseline TimeSformer~\cite{timesformer}, our video transformer backbone, and with Hyperformer~\cite{zhou2022hyperformer}, the 3D skeleton model used in 3D-SIM. We first observe the disparity in performance on Smarthome between Hyperformer and TimeSformer, owing this to the noisy poses in Smarthome and the importance of appearance in distinguishing the actions. This is further evident from the relatively small improvement seen on Smarthome when adding 3D poses to {\Large$\pi$}-ViT compared to NTU60 and NTU120. On the NTU datasets, we observe that the TimeSformer outperforms Hyperformer, but that both modalities are quite complementary (as evidenced by {\Large$\pi$}-ViT + 3D Poses).

\begin{figure}[h]
    \centering
    \includegraphics[width=\linewidth]{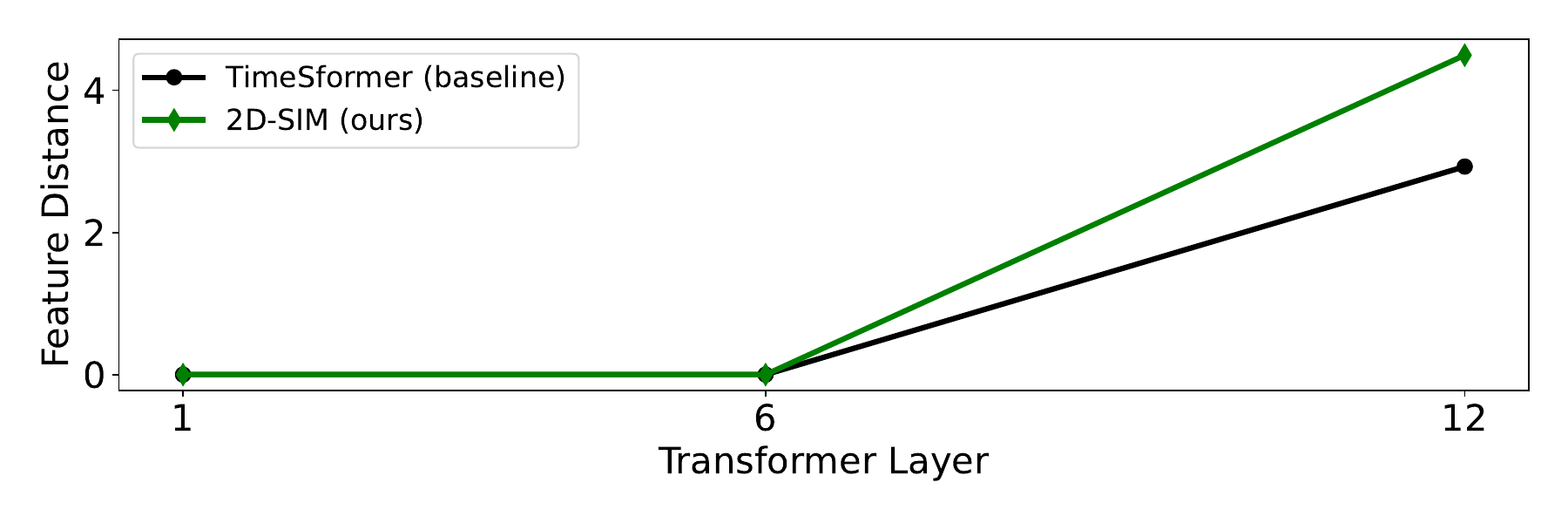}
    \caption{Average feature distance between pose tokens.}
    \label{fig:pose-pose-dist}
\end{figure}

\section{Feature Analysis of 2D-SIM}
Figure \ref{fig:pose-pose-dist} presents an investigation into the feature space of 2D-SIM, illustrating its enhanced capability in learning discriminative features for various human joints in comparison to a baseline model. This analysis was conducted by selecting a subset of videos from the Toyota-Smarthome dataset. The methodology involved computing the average distance of features between tokens corresponding to human joints across different layers within the video transformer. We find that towards the later layers of the video transformer, 2D-SIM is able to refine the representations to better disambiguate between the various human joints.

%%%%%%%%%%%%%%%%%%%%%%%%%%%%%%%%%%%%%%%%%%%%%%%%%%%%%%%%%%%%%%%%%%%%%%%%
% REFERENCES
%%%%%%%%%%%%%%%%%%%%%%%%%%%%%%%%%%%%%%%%%%%%%%%%%%%%%%%%%%%%%%%%%%%%%%%%
% \newpage
{
    \small
    \bibliographystyle{ieeenat_fullname}
    \bibliography{main}
}


\end{document}